\documentclass[conference]{IEEEtran}
\usepackage{times}
\usepackage{soul}
\usepackage[numbers]{natbib}
\usepackage{multicol}
\usepackage[bookmarks=true]{hyperref}
\usepackage[table,xcdraw,svgnames]{xcolor}
\usepackage{amsmath}
\usepackage{algorithm}
\usepackage{subcaption}
\usepackage{algorithmic}

\usepackage{eqparbox}

\usepackage{graphicx}
\usepackage{amsfonts}
\usepackage{multirow}
\let\labelindent\relax
\usepackage{enumitem}
\usepackage{bbm}
\usepackage{booktabs}
\usepackage{tikz}
\usepackage{highlight}
\usepackage{amsfonts}
\usepackage{multirow}
\usepackage{bbm}
\usepackage{hyperref}
\usepackage{siunitx}
\usepackage{float}
\usepackage{amssymb}
\DeclareMathOperator*{\argmax}{arg\,max}

\newtheorem{theorem}{Theorem}
\newtheorem{lemma}{Lemma}

\definecolor{start_col}{HTML}{ff7f00}
\definecolor{end_col}{HTML}{fff4ea}
\definecolor{mid_col}{HTML}{ffb977}

\newcommand\Mark[2][8.4]{%
  \rlap{\tikz[]{
        \shade[left color=start_col, right color=end_col, middle color=mid_col]
               (0,0) rectangle ++(#1*#2/100,0.3);}%
  }%
}

\begin{document}

\title{Learning Interpretable, High-Performing Policies for Autonomous Driving}


\author{\authorblockN{Rohan Paleja*, Yaru Niu*, Andrew Silva, Chace Ritchie, Sugju Choi, and Matthew Gombolay}
\authorblockA{Atlanta, Georgia 30332\\
Georgia Institute of Technology\\
$\{$rohan.paleja, yaruniu, andrew.silva, critchie7, schoi424, matthew.gombolay$\}$@gatech.edu}}


%

\maketitle
\begingroup\renewcommand\thefootnote{*}
\footnotetext{Authors contributed equally to this work.}
\endgroup
\begin{abstract}
\textcolor{black}{Gradient-based approaches in reinforcement learning (RL) have achieved tremendous success in learning policies for autonomous vehicles. While the performance of these approaches warrants real-world adoption, these policies lack interpretability, limiting deployability in the safety-critical and legally-regulated domain of autonomous driving (AD). 
AD requires interpretable and verifiable control policies that maintain high performance. We propose Interpretable Continuous Control Trees (ICCTs), a tree-based model that can be optimized via modern, gradient-based, RL approaches to produce high-performing, interpretable policies.
The key to our approach is a procedure for allowing direct optimization in a sparse decision-tree-like representation.
We validate ICCTs against baselines across six domains, showing that ICCTs are capable of learning interpretable policy representations that parity or outperform baselines by up to 33$\%$ in AD scenarios while achieving a $300$x-$600$x reduction in the number of policy parameters against deep learning baselines. Furthermore, we demonstrate the interpretability and utility of our ICCTs through a 14-car physical robot demonstration.}
\end{abstract}

\IEEEpeerreviewmaketitle

\section{Introduction}
\textcolor{black}{The deployment of autonomous vehicles (AVs) has the potential to increase traffic safety \cite{Katrakazas2015RealtimeMP}, decrease traffic congestion, increase average traffic speed in human-driven traffic \cite{Cui2021ScalableMD}, reduce $CO_2$ emissions, and allow for more affordable transportation \cite{Abe2019IntroducingAB}. Recent success in developing AVs with high-performance, real-time decision-making capabilities has been driven by the generation of continuous control policies produced via reinforcement learning (RL) with deep function approximators.}
However, while the performance of these controllers opens up the possibility of real-world adoption, the conventional deep-RL policies used in prior work \cite{Lillicrap2016ContinuousCW,Wu2017FlowAA,Cui2021ScalableMD} lack \emph{interpretability}, limiting deployability in safety-critical and legally-regulated domains \cite{doshi2017towards,letham2015interpretable,bhatt2019explainable,voigt2017eu}. 

White-box approaches, as opposed to typical black-box models (e.g., deep neural networks) used in deep-RL, model decision processes in a human-readable representation. Such approaches afford interpretability, allowing users to gain insight into the model's decision-making behavior. \textcolor{black}{In autonomous driving, such models would provide insurance companies, law enforcement, developers, and passengers with insight into how an autonomous vehicle (AV) reasons about state features and makes decisions.} Utilizing such white-box approaches within machine learning is necessary for the deployment of autonomous vehicles and essential in building trust, ensuring safety, and enabling developers to inspect and verify policies before deploying them to the real world \cite{olah2018building,hendricks2018generating,anne2018grounding}. In this work, we present a novel tree-based architecture that affords gradient-based optimization with modern RL techniques to produce high-performance, interpretable policies \textcolor{black}{for autonomous driving applications. We note that our proposed architecture can be applied to a multitude} of continuous control problems in robotics \cite{Lillicrap2016ContinuousCW}, protein folding \cite{Jumper2021HighlyAP}, and traffic regulation \cite{Cui2021ScalableMD}.

Prior work~\cite{olah2018building,Kim2015InteractiveAI,hendricks2018generating} has attempted to approximate interpretability via explainability, a practice that can have severe consequences \cite{Rudin2018StopEB}. 
While the explanations produced in prior work can help to partially explain the behavior of a control policy, the explanations are not guaranteed to be accurate or generally applicable across the state-space, leading to erroneous conclusions and a lack of accountability of predictive models \cite{Rudin2018StopEB}. \textcolor{black}{In autonomous driving, where understanding a decision-model is critical to avoiding collisions, local explanations are insufficient.} An \textit{interpretable model} provides a transparent \textit{global} representation of a policy's behavior. This model can be understood directly by its structure and parameters \cite{Ciravegna2021LogicEN} (e.g., linear models, decision trees, and our ICCTs), \textcolor{black}{offering verifiability} and guarantees that are not afforded by post-hoc explainability frameworks. Few works have attempted to learn an interpretable model directly; rather, prior work has attempted policy distillation to a decision tree \cite{Frosst2017DistillingAN,viper,wu2018beyond} or imitation learning via a decision tree across trajectories generated via a deep model \cite{Bastani2018VerifiableRL}, leaving much to be desired. Interpretable RL remains an open challenge \cite{Rudin2021InterpretableML}. In this work, we directly produce high-performance, interpretable policies represented by a minimalistic tree-based architecture augmented with low-fidelity linear controllers via RL, \textcolor{black}{providing a novel interpretable RL architecture}. Our Interpretable Continuous Control Trees are human-readable, \textcolor{black}{allow for closed-form verification (associated with safety guarantees)}, and parity or outperform baselines by up to $33\%$ in autonomous driving scenarios. In this work:
\begin{enumerate}[leftmargin=*]
    \item We propose Interpretable Continuous Control Trees (ICCTs), a novel tree-based model that can be optimized via gradient descent with modern RL algorithms to produce high-performance, interpretable continuous control policies. We provide several extensions to prior DDT frameworks to increase expressivity and allow for direct optimization on a sparse decision-tree-like representation.
    \item We empirically validate ICCTs across six continuous control domains, including four autonomous driving scenarios. 
    \item We provide a qualitative example of our ICCTs, displaying its interpretability and utility.
    \item \textcolor{black}{We demonstrate our algorithm with physical robots in a 14-car driving scenario and provide an online, easy-to-inspect visualization of the ego vehicle control policy.}
\end{enumerate}
\section{Related Work}
\label{sec:related_work}
\textcolor{black}{Due to recent accidents with autonomous vehicles (c.f.~\citet{Yurtsever2020ASO}), there has been growing interest in developing Explainable AI (xAI) approaches to understand an AV's decision-making and ensure robust and safe operation.} Explainable AI (xAI) is concerned with understanding and interpreting the behavior of AI systems \cite{Linardatos2021ExplainableAA}. In recent years, the necessity for human-understandable models has increased greatly for safety-critical and legally-regulated domains, many of which involve continuous control \textcolor{black}{(e.g., specifying joint torques for a robot arm or the steering angle for an autonomous vehicle)} \cite{Kim2017InterpretableLF,doshi2017towards}. 
In such domains, prior work \cite{Schulman2017ProximalPO,Lillicrap2016ContinuousCW,Fujimoto2018AddressingFA,haarnoja2018soft} has typically used highly-parameterized deep neural networks in order to learn high-performance policies, completely lacking in model transparency.

Interpretable machine learning approaches refers to a subset of xAI techniques that produce globally transparent policies (i.e., humans can inspect the entire model, as in a decision tree \cite{Breiman1983ClassificationAR,basak2004online,olaru2003complete} or rule list \cite{angelino2017learning,weiss1995rule,letham2015interpretable,chen2017optimization}). 
Decision trees \cite{Breiman1983ClassificationAR} represent a hierarchical structure where an input decision can be traced to an output via evaluation of decision nodes (i.e., ``test" on an attribute) until arrival at a leaf node. 
Decision nodes within the tree are able to split the problem space into meaningful subspaces, simplifying the problem as the tree gets deeper \cite{laptev2014convolutional,kontschieder2015deep,tanno2018adaptive}. 
Decision trees provide \textit{global} explanations of a decision-making policy that are valid throughout the input space \cite{Barbiero2021PyTorchEA}, as opposed to local explanations typically provided via ``post-hoc"  explainability techniques \cite{Ribeiro2019ModelAgnosticEA,Silva2021EncodingHD,Paleja2020InterpretableAP}. 
Several approaches have attempted to distill trained neural network models into decision trees \cite{Wu2020RegionalTR,viper}. \textcolor{black}{While these approaches produce interpretable models, the resulting model is an approximation of the neural network rather than a true representation of the underlying model. Our work, instead, directly learns an interpretable tree-based policy via reinforcement learning, producing a model that can be directly verified without utilizing error-prone post-hoc explainability techniques.
We emphasize that explainability stands in contrast to interpretability}, as explanations may fail to capture the true decision-making process of a model or may apply only to a local region of the decision-space, thereby preventing a human from building a clear or accurate mental model of the entire policy ~\cite{Rudin2018StopEB,lipton2018mythos,adadi2018peeking, Paleja_Utility_of_xAI}. 

Recently, \cite{Rudin2021InterpretableML} presented a set of grand challenges in interpretable machine learning to guide the field towards solving critical research problems that must be solved before machine learning can be safely deployed within the real world. 
In this work, we present a solution to directly assess two challenges: (1) Optimizing sparse logical models such as decision trees and (10) Interpretable reinforcement learning. We propose a novel high-performing, sparse tree-based architecture, Interpretable Continuous Control Trees (ICCTs), which allows end-users to directly inspect the decision-making policy and developers to verify the policy for safety guarantees. 




\section{Preliminaries}
In this section, we review differentiable decision trees (DDTs)
and reinforcement learning. 
\begin{figure*}[t]
    \centering
    \includegraphics[width=\textwidth]{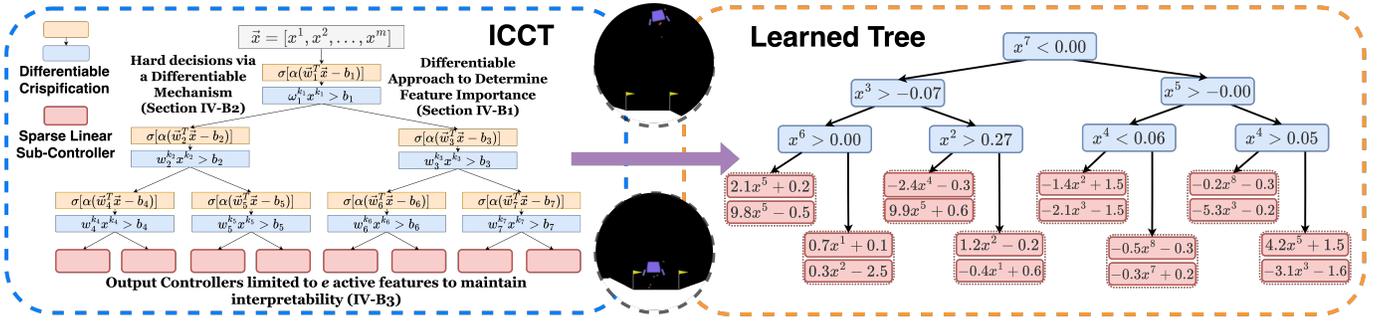}
    \caption{\textcolor{black}{The ICCT framework (left) displays decision nodes, both in their fuzzy form (orange blocks) and crisp form (blue blocks\protect\footnotemark), and sparse linear leaf controllers with pointers to sections discussing our contributions. A learned representation of a high-performing ICCT policy in Lunar Lander (right) displays the interpretability of our ICCTs. Each decision node is conditioned upon only a single feature and the sparse linear controllers (to control the main engine throttle and left/right thrusters) are set to have only \textbf{one} active feature.}}
    
    \label{fig:dt}
\end{figure*}

\subsection{Differentiable Decision Trees (DDTs)}
\label{sec:ddts}
\textcolor{black}{Prior work has proposed differentiable decision trees (DDTs)  \cite{suarez1999globally,Silva2021EncodingHD} -- a neural network architecture that takes the topology of a decision tree (DT). Similar to a decision tree, DDTs contain decision nodes and leaf nodes; however, each decision node within the DDT utilizes a sigmoid activation function (i.e., a ``soft" decision) instead of a Boolean decision (i.e., a ``hard" decision). Each decision node, $i$, is represented by a sigmoid function, displayed in Equation \ref{eq:ddt_decision_node}.}
\begin{equation}
\label{eq:ddt_decision_node}
y_i = \frac{1}{1+\exp(-\alpha(\vec{w}_{i}^{T} \vec{x} - b_i))}
\end{equation}
\textcolor{black}{Here, the features vectors describing the current state, $\vec{x}$, are weighted by $\vec{w}_i$, and a splitting criterion, $b_i$, is subtracted to form the splitting rule. $y_i$ is the probability of decision node $i$ evaluating to \textsc{True}, and $\alpha$ governs the steepness of the sigmoid activation, where $\alpha \rightarrow \infty$ results in a step function.} Prior work with discrete-action DDTs modeled each leaf node with a probability distribution over possible output classes \cite{Silva2021EncodingHD, Paleja2020InterpretableAP}. Leaf node distributions are then weighted by the probability of reaching the respective leaf and summed to produce a final action distribution over possible outputs.

\subsubsection{Conversion of a DDT to a DT}
\label{sec:prolo_crisp}
DDTs with decision nodes represented in the form of Equation \ref{eq:ddt_decision_node} are not interpretable.
As DDTs maintain a one-to-one correspondence to DTs with respect to their structure, prior work \cite{Silva2021EncodingHD, Paleja2020InterpretableAP} proposed methodology to convert a DDT into an interpretable decision tree (a process termed ``crispification").
To create an interpretable, ``crisp" tree from a differentiable form of the tree, prior work adopted a simplistic procedure. Starting with the differentiable form, prior work first converts each decision node from a linear combination of all variables into a single feature check (i.e., a 2-arity predicate with a variable and a threshold). The feature reduction is accomplished by considering the feature dimension corresponding to the weight with the largest magnitude \textcolor{black}{(i.e., most impactful)}, $ k = \argmax_{j}|w_{i}^{j}|$ \textcolor{black}{where $j$ represents the feature dimension},
resulting in the decision node representation $y_i = \sigma(\alpha(w^k_ix^k-b_i))$.
The sigmoid steepness, $\alpha$, is also set to infinity, resulting in a ``hard" decision (branch left OR right) \cite{Silva2021EncodingHD,Paleja2020InterpretableAP}. After applying this procedure to each decision node, decision nodes are represented by  $y_i = \mathbbm{1}(w^k_ix^k-b_i > 0)$.
As each leaf node is represented as a probability mass function over output classes in prior work, leaf nodes, $l$, must be modified to produce a single output class, $o$, during crispification. As such, we can apply an argument max, $o_d = \argmax_a l^a_d$, \textcolor{black}{where $a$ denotes the action dimension,} to find the maximum valued class within the $d^{th}$ leaf distribution.

\noindent\textbf{Drawbacks:} This simplistic crispification procedure results in an interpretable crisp tree that is inconsistent with the original DDT (model differences arise from each \textit{argmax} operation \textcolor{black}{and setting $\alpha$ to infinity}). These inconsistencies can lead to performance degradation of the interpretable model, as we show in Section \ref{sec:results}, and results in an interpretable model that is not representative of and inconsistent with the model learned via reinforcement learning.\footnotetext{For figure simplicity, when displaying the crisp node (blue block), we assume $\alpha>0$ in the fuzzy node (orange block). If $\alpha<0$, the sign of the inequality would be flipped (i.e., $w_i^{k_i} x^{k_i}<b$).} 

\textcolor{black}{In our work, we design a novel architecture that updates its parameters via gradient descent while maintaining an interpretable decision-tree-like representation, thereby avoiding any inconsistencies generated through a \textcolor{black}{post-training} crispification procedure.
To the best of our knowledge, we are the first work to deploy an interpretable tree-based framework for continuous control.}

\subsection{Reinforcement Learning (RL)}
A Markov Decision Process (MDP) $M$ is defined as a 6-tuple $\langle S,A,R,T,\gamma,\rho_0\rangle$. $S$ is the state-space, $A$ is the action-space, $R(s,a)$ is the reward received by an agent for executing action, $a$, in state, $s$, $T(s^\prime|s,a)$ is the probability of transitioning from state, $s$, to state, $s'$, when applying action, $a$, $\gamma\in [0,1]$ is the discount factor, and $\rho_0(s)$ is the initial state distribution. 
A policy, $\pi(a|s)$, gives the probability of an agent taking action, $a$, in state, $s$. 
The goal of RL is to find the optimal policy, 
$\pi^*=\arg\max_\pi \mathbb{E}_{\tau\sim\pi}\left[\sum_{t=0}^T{\gamma^tR(s_t,a_t)}\right]$ to maximize cumulative discounted reward, where $\tau=\langle s_0,a_0,\cdots,s_T,a_T\rangle$ is an agent's trajectory. 
In this work, while ICCTs are framework-agnotistic (i.e., ICCTs will work with any update rule), we proceed with Soft Actor-Critic (SAC) \cite{haarnoja2018soft} as our RL algorithm \textcolor{black}{due to its learning stability and high sample efficiency}. The actor objective within SAC is given in Equation \ref{eqn:sac-actor}, where $Q_w(s_t,a_t)$ is expected, future discounted reward as parameterized by $\omega$ \textcolor{black}{and $\alpha_{\pi}$ is a weighting parameter to maximize the entropy of the stochastic policy.}
\begin{align}
    \label{eqn:sac-actor}
        J_\pi(\theta) = \mathbb{E}_{s_t \sim\mathcal{D}}[\mathbb{E}_{a_t \sim\pi_\theta}[&\alpha_{\pi} \log(\pi_\theta(a_t|s_t)) -  Q_\omega(s_t, a_t)]]
\end{align}

\section{Method}
\label{sec:method}
\textcolor{black}{In this section, we introduce our ICCTs, a novel interpretable reinforcement learning architecture. ICCTs are able to maintain interpretability while representing high-performance continuous control policies, \textcolor{black}{making them suitable for applications that require trust and accountability such as autonomous vehicle control}. We provide several extensions to prior DDT frameworks within our proposed architecture including 1) a differentiable crispification procedure allowing for optimization in a sparse decision-tree like representation, and 2) the addition of sparse linear leaf controllers to increase expressivity while maintaining legibility. 
}

\subsection{ICCT Architecture}
\label{sec:icct_arch}
Our ICCTs are initialized to be a symmetric decision tree with \textcolor{black}{$n_l$ decision leaves (red nodes in Figure \ref{fig:dt}) and $n_l-1$ decision nodes (blue nodes in Figure \ref{fig:dt}). The tree depth is given by $\log_2(n_l)$}. Each decision leaf is represented by a linear sparse controller that is \textcolor{black}{operated on} $\vec{x}$. Decisions are routed via decision nodes towards a leaf controller, which is then used to produce the continuous control output (e.g., acceleration or steering wheel angle).
Our ICCT is similar to hierarchical models which maintain a high-level controller over several low-level controllers. Prior work has shown this to be a successful paradigm in continuous control \cite{Nachum2018DataEfficientHR}.

Each decision node, $i$, has an activation steepness weight, \textcolor{black}{$\alpha$}, associated weights, $\vec{w}_i$, with cardinality, $m$, matching that of the input feature vector, $\vec{x}$, and a scalar bias term, $b_i$, similar to that of Equation \ref{eq:ddt_decision_node}. 
\textcolor{black}{Each leaf node, $l_d$, where \textcolor{black}{$d \in \{1,\dots,n_l\}$}, contains per-leaf weights, $\vec{\beta}_d \in \mathbb{R}^{m}$, per-leaf selector weights that learn the relative importance of \textcolor{black}{candidate features}, $\vec{\theta}_d \in \mathbb{R}^{m}$, per-leaf bias terms, $\vec{\phi}_{d} \in \mathbb{R}^{m}$, and per-leaf scalar standard deviations, $\gamma_d$.} 
We note that if the action space is multi-dimensional, then \emph{only} the leaf controllers (and associated weights) are expanded across $|A|$ dimensions, where $|A|$ is the cardinality of the action space. For each action dimension, the mean of the output action distribution is represented by the linear controller\textcolor{black}{, $l_d$.
\begin{equation}
    \label{eq:complete_leaf}
    l_d \triangleq (\vec{u}\circ\vec{\beta}_d)^T (\vec{u}\circ\vec{x}) - \vec{u}^T\vec{\phi}_d
\end{equation}
Before enforcing leaf node sparsity (Section~\ref{sec:sparse_sub_controllers}), $\vec{u}=[1, ..., 1]^T$ is an all-ones vector, representing a selection of all input features for the leaf node, in which case Equation~\ref{eq:complete_leaf} can be simplified as $l_d=\vec{\beta}_d^T \vec{x} - \vec{u}^T\vec{\phi}_d$. The output action can be determined via sampling ($a\sim \mathcal{N}(\vec{\beta}_d^T \vec{x} - \vec{u}^T\vec{\phi}_d, \gamma_d)$) during training and directly via the mean during runtime. We term decision nodes that are represented as Equation \ref{eq:ddt_decision_node} as fuzzy decision nodes, displayed by the orange rectangles within the left-hand side of Figure \ref{fig:dt}. Similarly, we term the leaf node, $l_d$, which is represented in the dense representation of $\vec{\beta}_d^T \vec{x} - \vec{u}^T\vec{\phi}_d$, as a fuzzy leaf node. It is worth noting that we parameterize the bias term as a vector $\vec{\phi}_d$ instead of a scalar to provide a corresponding bias for each feature and facilitate feature-wise optimization for the bias.}

Utilizing a novel differentiable crispification procedure to convert fuzzy decision nodes into crisp decision nodes (i.e., 2-arity predicate with a variable and a threshold) and fuzzy leaf nodes into sparse leaf nodes (i.e., linear controller conditioned upon a subset of features), our model representation follows that of a decision tree with sparse linear controllers at the leafs (shown on the right-hand side of Figure \ref{fig:dt}). We further discuss our differentiable crispification procedure in Sections \ref{sec:decision_nodes_crispification} and \ref{sec:decision_outcome_crispification} (i.e., the mechanism that translates orange blocks to blue within Figure \ref{fig:dt}) \textcolor{black}{and leaf controller sparsification procedure in Section \ref{sec:sparse_sub_controllers}.}

While decision trees (DT) are generally considered interpretable \cite{letham2015interpretable}, trees of arbitrarily large depths can be difficult to understand \cite{Ghose2020InterpretabilityWA} and simulate \cite{lipton2018mythos}. A sufficiently sparse DT is desirable and considered interpretable \cite{lakkaraju}. Utilizing linear controllers at the leaves also allows us to maintain interpretability, as linear controllers are widely used and generally considered interpretable for humans \cite{Hein2020InterpretableCB}. However, similarly, for large feature spaces typically encountered in real-world problems, such a controller would not be interpretable. \textcolor{black}{As such, in our work, we utilize \emph{sparse} linear controllers at the leaves to balance the trade-off between sparsity/complexity in logic, model depth, and performance.} 


\subsection{\textcolor{black}{ICCT Key Elements}}
In this section, we discuss our ICCT's interpretable procedure for determining an action given an input feature. As our ICCT configuration maintains interpretability both during training via RL and deployment, the inference of an action \textcolor{black}{must allow gradient flow. We present a novel approach that allows for direct optimization of sparse logical models via an online differentiable crispification procedure to determine feature importance (Section \ref{sec:decision_nodes_crispification}) and allows for bifurcate decisions (Section \ref{sec:decision_outcome_crispification}).} In Algorithm \ref{alg:algorithm_training}, we provide general pseudocode representing our ICCT's decision-making process. 

At each timestep, the ICCT model, $I(\cdot)$, receives a state feature, $\vec{x}$. To determine an action in an interpretable form, in Steps 1 and 2 of Algorithm \ref{alg:algorithm_training}, we start by applying the differentiable crispification approaches of \textsc{Node$\_$Crisp} and \textsc{Outcome$\_$Crisp} to decision nodes so that each decision node is only conditioned upon a single variable (Section \ref{sec:decision_nodes_crispification}), and the evaluation of a decision node results in a Boolean (Section \ref{sec:decision_outcome_crispification}). Once the operations are completed, in Step 3, we can utilize the input feature, $\vec{x}$, and logically evaluate each decision node until arrival at a linear leaf controller (\textsc{Interpretable$\_$Node$\_$Routing}). The linear leaf controller is then modified, in Step 4, to only condition upon $e$ features, where $e$ is a sparsity parameter specified a priori (Section \ref{sec:sparse_sub_controllers}). Finally, an action can be determined via sampling from a Gaussian distribution conditioned upon the mean generated via the input-parameterized sparse leaf controller, $l^*_d$, and scalar variance maintained within the leaf, $\gamma_d$, (Step 6) during training or directly through the outputted mean (Step 8) during runtime.

\begin{algorithm}[h]
\caption{ICCT Action Determination}
\label{alg:algorithm_training}
\textbf{Input}: \small ICCT $I(\cdot)$, state feature $\vec{x} \in S$, controller sparsity $e$, training flag \textcolor{black}{$t \in \{\textsc{True, False}\}$} \\
\textbf{Output}: action $a \in \mathbb{R}$ 
\begin{algorithmic}[1] 
\STATE \textsc{Node$\_$Crisp}($\sigma(\alpha(\vec{w}^T_i \vec{x} - b_i))$)
$\rightarrow \sigma(\alpha(w_i^k x^k - b_i))$
\STATE \textsc{Outcome$\_$Crisp}($\sigma(\alpha(w_i^k x^k - b_i))$)
$\rightarrow \mathbbm{1}(\alpha(w_i^k x^k - b_i)>0)$
\STATE $l_d \leftarrow$ \textsc{Interpretable$\_$Node$\_$Routing}($\vec{x}$)
\STATE $l_d^* \leftarrow$ \textsc{Enforce$\_$Controller$\_$Sparsity}($e$, $l_d$)
\IF {$t$}
\STATE \textcolor{black}{$a\sim \mathcal{N}(l_d^*(\vec{x}), \gamma_d)$}
\ELSE
\STATE $a\leftarrow l_d^*(\vec{x})$
\ENDIF 
\end{algorithmic}
\end{algorithm}

\subsubsection{Decision Node Crispification}
\label{sec:decision_nodes_crispification}
The \textsc{Node$\_$Crisp} procedure in Algorithm \ref{alg:algorithm_training} \textcolor{black}{recasts} each decision node to split upon a single dimension of $\vec{x}$ \textcolor{black}{to achieve sparsity while maintaining differentiable}.
Instead of using a non-differentiable argument max function as in \citet{Silva2021EncodingHD} to determine the most impactful feature dimension, we utilize a softmax function, also known as softargmax \cite{Goodfellow-et-al-2016}, described by Equation \ref{eq:softmax}. In this equation, we denote the softmax function as $f(\cdot)$, which takes as input a set of class weights and produces class probabilities. \textcolor{black}{Here, $\vec{w}_i$ represents a categorical distribution with class weights, individually denoted by $w_i^j$}, and $\tau$ is the temperature, determining the steepness of $f(\cdot)$. 
\textcolor{black}{
\begin{equation}
\label{eq:softmax}
    f(\vec{w}_i)_k = \frac{\exp{\big(\frac{w_i^k}{\tau}\big)}}{\sum_j^m \exp{\big(\frac{w_i^j}{\tau}\big)}} 
\end{equation}
}

\noindent While setting the temperature near-zero would satisfy our objective of producing a one-hot vector, where the outputted class probability of the index of the most impactful feature would be one, this operation can lead to \textcolor{black}{large gradient variance and unstable training}. We therefore set $\tau$ equal to 1 \textcolor{black}{which we find effective empirically}, and utilize a \textcolor{black}{differentiable $\text{one\_hot}$ function, $g(\cdot)$,} which produces a one-hot vector with the element associated with the highest-weighted class set to one and all other elements set to zero. We display the differentiable procedure for determining the weight associated with the largest magnitude in Equation \ref{eq:crispification}. 
\begin{equation}
\label{eq:crispification}
    \vec{z_i} = g(f(|\vec{w_i}|))
\end{equation}
Here, $|\vec{w_i}|$ represents a vector with absolute elements within $\vec{w_i}$. \textcolor{black}{The one-hot encoding $\vec{z_i}$ can be element-wise multiplied by the original weights to produce a new set of weights with only one active weight, $\vec{z}_i\circ \vec{w}_i\rightarrow\vec{w}_i'$. Accordingly, the decision node representation is transferred from $\sigma(\alpha(\vec{w}^T_i \vec{x} - b_i)) \to  \sigma(\alpha(\vec{w}_{i}'^{T} \vec{x} - b_i))=\sigma(\alpha(w_i^k x^k - b_i))$, where $k$ is the index of the most impactful feature. \emph{We maintain differentiability in the procedure described in Equation \ref{eq:crispification} by utilizing the straight-through trick \cite{Bengio2013EstimatingOP}.} The differentiable one-hot operation is further described in Algorithm \ref{alg:argmax} within the Appendix (Section \ref{sec:diff_argmax}). 
We provide an algorithm detailing the \textsc{Node$\_$Crisp} procedure within the Appendix (Section \ref{sec:append_node_crisp}).
Below, we conduct a short example detailing our procedure.}

\textit{Example:} \textcolor{black}{Consider} we have a two-leaf decision tree (one decision node) with an input feature, \textcolor{black}{$\vec{x}=[2,3]^T$} with a cardinality of 2 (i.e. $m = 2$), associated weights of $\vec{w_1} = [2,1]^T$, and a bias term $b_1$ of 1. The sigmoid steepness, $\alpha$ is also set equal to 1 for simplicity. It is easily seen that the most impactful weight within the decision node is $w_1^1=2$. Utilizing Equation \ref{eq:crispification}, we can compute \textcolor{black}{$\vec{z_1}=[1,0]^T$}. Multiplying $\vec{z_1}$ to the original weights, $\vec{w_1}$ and input feature, $\vec{x}$, subtracting $b_1$, and scaling by $\alpha$, we have an crisp decision node $\sigma(2x^1-1)$ or $\sigma(3)=0.95$. Here, $0.95$ is the probability that the decision node evaluates to \textcolor{black}{\textsc{True}}. \textcolor{black}{We display a depiction of this example in the left-hand side of Figure \ref{fig:example_diff}.}

\subsubsection{Decision Outcome Crispification}
\label{sec:decision_outcome_crispification}
Here, we describe the second piece of our online differentiable crispification procedure, noted as \textsc{Outcome$\_$Crisp} in Algorithm \ref{alg:algorithm_training}.
\textsc{Outcome$\_$Crisp} translates the outcome of a decision node so that the outcome is a Boolean decision rather than a set of probabilities generated via a sigmoid function (i.e., $p =y_i$ for True/Left Branch and $q=1-y_i$ for False/Right Branch). We start by creating a soft vector representation of the decision node output $\vec{v}_i = [\alpha(w_i^k x^k - b_i), 0]$, for the $i^{th}$ decision node. Placing $\vec{v}_i$ through a softmax operation, we can produce the probability of tracing down the left branch or right.
We can then apply \textcolor{black}{the differentiable $\text{one-hot}$ function, $g(\cdot)$}
to produce a hard decision of whether to branch left or right, denoted by $y_i$ and described by Equation \ref{eq:decision_outcome_crispfication}. Essentially, the decision node will evaluate to \textcolor{black}{\textsc{True}} if $\alpha(w_i^k x^k - b_i)>0$ and right otherwise. This process can be expressed as an indicator function $ \mathbbm{1}(\alpha(w_i^k x^k - b_i)>0)$. 
\begin{equation}
    \label{eq:decision_outcome_crispfication}
    (y_i,1-y_i) = g(f(\vec{v}_i)) 
\end{equation}
We note the procedure of $g(f(\vec{v}_i))$ is highly similar to that in Equation \ref{eq:crispification}, both outputting a one-hot vector, with the former input being the decision node weights, $|\vec{w}_i|$, and the latter input being the soft vector representation of the decision node outcome, $\vec{v}_i$.
We provide an algorithm detailing the \textsc{Outcome$\_$Crisp} procedure within the Appendix (Section \ref{sec:append_outcome_crisp}), including the operations to maintain gradients via the straight-through trick (Algorithm \ref{alg:argmax}).

\begin{figure*}[ht]
\centering
\includegraphics[width=0.85\textwidth]{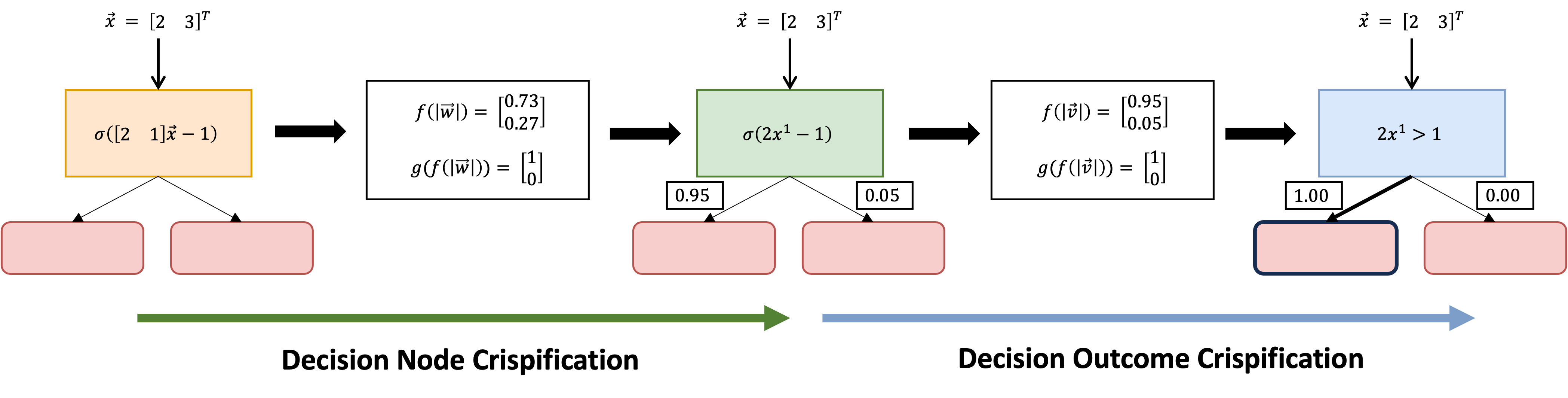}
\caption{\textcolor{black}{This figure displays the process of decision node crispification and decision outcome crispification across the Examples within Section \ref{sec:decision_nodes_crispification} and Section \ref{sec:decision_outcome_crispification}.}}
\label{fig:example_diff}
\end{figure*}

\textit{Example: Continuing the example in Section \ref{sec:decision_nodes_crispification}}, we can take the outputted crisp decision node and generate a vector \textcolor{black}{$\vec{v}_1 =[2x^1-1,0]^T$}, or by substituting the input feature, \textcolor{black}{$\vec{v}_1 =[3,0]^T$}. Performing the operations specified in Equation \ref{eq:decision_outcome_crispfication}, we receive the intermediate output from the softmax \textcolor{black}{$[0.95, 0.05]^T$} (rounded to two decimal places), the one-hot vector \textcolor{black}{$[1,0]^T$} after performing the $\text{one\_hot}$ operation, and finally $y_1 = 1$, denoting that the decision-tree should follow the left branch. \textcolor{black}{We display a depiction of this example on the right-hand side of Figure \ref{fig:example_diff}.}

\textit{Conversion to a Simple Form:}
The above crispification process produces decision-tracing equal to that of a DT. The node representation can be further simplified to that of Figure \ref{fig:dt} by algebraically reducing each crisp decision node to $ x^k>\frac{b_i}{w_i^k}$ (given $\alpha w_i^k>0$) or $ x^k<\frac{b_i}{w_i^k}$ (given $\alpha w_i^k<0$).

\subsubsection{Sparse Linear Leaf Controllers}
\label{sec:sparse_sub_controllers}
After applying the decision node and outcome crispification to all decision nodes and outcomes, the decision can be routed to leaf node (Step 3). This section describes the procedure to translate a linear leaf controller to condition upon $e$ features (\textsc{Enforce$\_$Controller$\_$Sparsity} procedure in Algorithm \ref{alg:algorithm_training}), enforcing sparsity within the leaf controller and thereby, enhancing ICCT interpretability. As noted in Section \ref{sec:icct_arch}, our ability to utilize sparse sub-controllers allows us to balance the interpretability-performance tradeoff. The sparsity of the linear sub-controllers ranges from setting $e=0$ and maintaining static leaf distributions, where each leaf node contains scalar value representing the mean (i.e., ICCT-static in Section \ref{sec:results}), to $e=m$, containing a linear controller parameterized by the entire feature space of $\vec{x}$ (i.e., ICCT-complete in Section \ref{sec:results}).
Equation \ref{eq:sparse_sub_impact} displays the procedure for determining a $\text{k\_hot}$ encoding, $\vec{u}_d$, that represents the $k$ \textcolor{black}{(or in our case, $e$)} most impactful selection weights within a leaf's linear controller. The $\text{k\_hot}$ function, denoted by $h(\cdot)$, takes as input a vector of class weights and returns an equal-dimensional vector with $k$ elements set to one. The indexes associated with the elements set to one match the $k$ highest-weighted elements within the input feature.
\begin{equation}
    \label{eq:sparse_sub_impact}
    \vec{u}_d = h (f(|\vec{\theta}_d|))
\end{equation}

\noindent Here, $|\vec{\theta}_d|$ represents a vector with absolute elements within $\vec{\theta}_d$. Similarly, we maintain differentiability \textcolor{black}{and formulate a differentiable top-$k$ function} in Equation \ref{eq:sparse_sub_impact} by utilizing the straight-through trick \citep{Bengio2013EstimatingOP} \textcolor{black}{and iteratively applying \textsc{diff\_argmax}$(\cdot)$ (Algorithm \ref{alg:argmax}) for $k$ times}.
\textcolor{black}{In Equation \ref{eq:translate_to_sparse_submodel}, we transform a fuzzy leaf node, for leaf, $l_d$ \textcolor{black}{(represented as Equation \ref{eq:complete_leaf}), into a sparse linear sub-controller, $l_d^*$, with the sparse feature selection vector, $\vec{u_d}$, given by Equation~\ref{eq:sparse_sub_impact}.}}
\begin{equation}
    \label{eq:translate_to_sparse_submodel}
    l_d^* \triangleq (\vec{u}_d \circ \vec{\beta}_d)^T (\vec{u}_d \circ \vec{x}) + \vec{u}_d^T \vec{\phi}_{d}
\end{equation}
A depiction of the sparse sub-models can be seen at the bottom of Figure \ref{fig:dt}, where the sparsity of the sub-controllers, $e$, is set to 1 and the dimension of the action space is 2.

\textcolor{black}{\noindent \textbf{Summary:} In this section, we discuss our novel interpretable reinforcement learning architecture, ICCTs. We present a description of components of ICCTs, including decision nodes and linear leaf controllers, and provide a differentiable crispification procedure allowing for optimization in a sparse decision-tree like representation. To the best of our knowledge, we present the first truly interpretable tree-based framework for continuous control.}

\begin{table*}
\caption{\textcolor{black}{In this table, we display the results of our evaluation. 
For each evaluation, we report the mean ($\pm$ standard error) and the complexity of the model required to generate such a result. Our table is broken into three segments, the first containing equally interpretable approaches that utilize static distributions at their leaves. The second segment contains interpretable approaches that maintain linear controllers at their leaves. The ordering of methods denotes the relative interpretability. The third segments displays black-box approaches. We bold the highest-performing method in each segment, and break ties in performance by model complexity. \textcolor{black}{We color table elements in association with the number of parameters and performance. 
Reddish colors relate to a larger number of policy parameters and lower average reward.}}}
\resizebox{\textwidth}{!}{%
\begin{tabular}{@{}ccccccc@{}}
\multicolumn{1}{c}{\large Worst to Best:} & \multicolumn{6}{l}{\Mark{230}} \\
\specialrule{.2em}{.1em}{.1em} 
\multirow{2}{*}{Method} & \multicolumn{2}{c|}{Common Continuous Control Problems} & \multicolumn{4}{c}{\textcolor{black}{Autonomous Driving Problems}} \\ \cmidrule(l){2-7} 
& Inverted Pendulum & \multicolumn{1}{c|}{Lunar Lander} & Lane Keeping & Single-Lane Ring & Multi-Lane Ring & Figure-8\\ \specialrule{.2em}{.1em}{.1em} \multirow{2}{*}{DT} & \cellcolor[HTML]{ff9023}$155.0\pm0.9$& \cellcolor[HTML]{ff9833}$-285.5\pm15.6$  &\cellcolor[HTML]{ff8f21}$-359.0 \pm 11.0$ &  \cellcolor[HTML]{fff4ea}\textbf{123.2 $\pm$ 0.03} &\cellcolor[HTML]{ff7f00}$503.2\pm24.8$ & \cellcolor[HTML]{ffcc9d}$831.1\pm1.1$  \\   & \cellcolor[HTML]{ffb875}256 leaves (766 params)  & \cellcolor[HTML]{ffb875}256 leaves (1022 params) & \cellcolor[HTML]{ffb977}256 leaves (766 params) &  \cellcolor[HTML]{fff3e8}32 leaves (94 params)  & \cellcolor[HTML]{ffdec0}256 leaves (1022 params) & \cellcolor[HTML]{ffc996}256 leaves (766 params)  \\  \midrule \multirow{2}{*}{DT w$\backslash$ DAgger} & \cellcolor[HTML]{ffd8b4}$776.6\pm54.2$& \cellcolor[HTML]{ffe1c5}$184.7\pm17.3$  &\cellcolor[HTML]{ffe9d5}\textbf{395.2 $\pm$ 13.8} & \cellcolor[HTML]{fff0e3} 121.5 $\pm$ 0.01 &\cellcolor[HTML]{ffeede}$1249.4\pm3.4$ &  \cellcolor[HTML]{fff4ea}\textbf{1113.8 $\pm$ 9.5}  \\   & \cellcolor[HTML]{ffeede}32 leaves (94 params)  & \cellcolor[HTML]{ffefe1}32 leaves (126 params) & \cellcolor[HTML]{fff4ea}16 leaves (46 params) &  \cellcolor[HTML]{fff4ea}16 leaves (46 params)  &  \cellcolor[HTML]{fff4ea}31 leaves (122 params) &  \cellcolor[HTML]{fff4ea}16 leaves (46 params)  \\  \midrule   \multirow{2}{*}{CDDT-Crisp} & \cellcolor[HTML]{ff7f00}$5.0\pm0.0$ & \cellcolor[HTML]{ff7f00}$-451.6\pm97.3$ &\cellcolor[HTML]{ff7f00}$-43526.0\pm15905.0$ &\cellcolor[HTML]{ff7f00}$68.1\pm18.7$ &\cellcolor[HTML]{ff962f}$664.5\pm192.6$ & \cellcolor[HTML]{ff8813}$322.9\pm47.1$  \\   & \cellcolor[HTML]{fff4ea}2 leaves (5 params) & \cellcolor[HTML]{fff4ea}8 leaves (37 params) & \cellcolor[HTML]{fff3e8}16 leaves (61 params) & \cellcolor[HTML]{fff4ea}16 leaves (61 params) & \cellcolor[HTML]{fff4ea}16 leaves (77 params) &  \cellcolor[HTML]{fff4ea}16 leaves (61 params)  \\  \midrule  \multirow{2}{*}{ICCT-static} & \cellcolor[HTML]{fff2e5}\textbf{984.0 $\pm$ 10.4} & \cellcolor[HTML]{ffe2c7}\textbf{192.4 $\pm$ 10.7} & \cellcolor[HTML]{ffe7d0}374.2$\pm$55.8 &\cellcolor[HTML]{ffeddc}$120.5\pm0.5$ & \cellcolor[HTML]{fff2e5}\textbf{1271.7 $\pm$ 4.1} & \cellcolor[HTML]{ffe4cc} 1003.8$\pm$27.2  \\   & \cellcolor[HTML]{ffead7}32 leaves (125 params) & \cellcolor[HTML]{ffeddc}32 leaves (157 params) & \cellcolor[HTML]{fff3e8}16 leaves (61 params) & \cellcolor[HTML]{fff4ea}16 leaves (61 params) & \cellcolor[HTML]{fff4ea}16 leaves (77 params) & \cellcolor[HTML]{fff4ea}16 leaves (61 params) \\  \specialrule{.2em}{.1em}{.1em}   \multirow{2}{*}{ICCT-1-feature} & \cellcolor[HTML]{fff4ea}$1000.0\pm0.0$ & \cellcolor[HTML]{ffe2c7}$190.1\pm13.7$ &\cellcolor[HTML]{ffeede}$437.6\pm7.0$ &\cellcolor[HTML]{fff0e3}$121.6\pm0.5$ &\cellcolor[HTML]{fff2e5}$1269.6\pm10.7$ &\cellcolor[HTML]{ffeede}$1072.4\pm37.1$   \\   & \cellcolor[HTML]{fff2e5}8 leaves (45 params) & \cellcolor[HTML]{fff3e8}8 leaves (69 params) & \cellcolor[HTML]{fff0e3}16 leaves (93 params) & \cellcolor[HTML]{fff3e8}16 leaves (93 params) & \cellcolor[HTML]{fff4ea}16 leaves (141 params) & \cellcolor[HTML]{fff2e5}16 leaves (93 params)    \\  \midrule  \multirow{2}{*}{ICCT-2-feature} & \cellcolor[HTML]{fff4ea}$1000.0\pm0.0$ & \cellcolor[HTML]{ffeddc}$258.4\pm7.0$ &\cellcolor[HTML]{fff0e3}$458.5\pm6.3$ & \cellcolor[HTML]{fff0e3}\textbf{121.9 $\pm$ 0.5} & \cellcolor[HTML]{fff4ea}1280.4$\pm$7.3 & \cellcolor[HTML]{fff0e3} \textbf{1088.6 $\pm$ 21.6}  \\   & \cellcolor[HTML]{fff3e8}4 leaves (29 params) & \cellcolor[HTML]{fff0e3}8 leaves  (101 params) & \cellcolor[HTML]{ffeede} 16 leaves (125 params) & \cellcolor[HTML]{fff2e5}16 leaves (125 params) & \cellcolor[HTML]{fff3e8}16 leaves (205 params) &  \cellcolor[HTML]{ffefe1}16 leaves (125 params) \\  \midrule  \multirow{2}{*}{ICCT-3-feature} & \cellcolor[HTML]{fff4ea}\textbf{1000.0 $\pm$ 0.0} & \cellcolor[HTML]{ffefe1}$275.8\pm1.5$ &\cellcolor[HTML]{ffefe1}$448.8\pm3.0$ &\cellcolor[HTML]{ffeddc}$120.8\pm0.5$ &\cellcolor[HTML]{fff4ea}\textbf{1280.8 $\pm$ 7.7} &\cellcolor[HTML]{ffead7}$1048.7\pm46.7$   \\   & \cellcolor[HTML]{fff4ea}2 leaves (17 params) & \cellcolor[HTML]{ffefe1}8 leaves (133 params) & \cellcolor[HTML]{ffebda}16 leaves (157 params) & \cellcolor[HTML]{ffefe1}16 leaves (157 params) & \cellcolor[HTML]{fff0e3}16 leaves (269 params) &  \cellcolor[HTML]{ffeede}16 leaves (157 params)  \\   \midrule  \multirow{2}{*}{ICCT-L1-sparse} & \cellcolor[HTML]{fff4ea}$1000.0\pm0.0$ & \cellcolor[HTML]{ffeede}$265.2\pm4.3$ &\cellcolor[HTML]{fff2e5}$465.5\pm4.3$ & \cellcolor[HTML]{fff0e3}$121.5\pm0.3$ & \cellcolor[HTML]{fff3e8}$1275.3\pm6.7$ & \cellcolor[HTML]{ffe3c9} $993.2\pm14.6$  \\   & \cellcolor[HTML]{fff3e8}4 leaves (29 params) & \cellcolor[HTML]{ffeddc}8 leaves (165 params) & \cellcolor[HTML]{ffe3c9}16 leaves (253 params) & \cellcolor[HTML]{ffd5ad}16 leaves (765 params) & \cellcolor[HTML]{ffc288}16 leaves (2189 params) &  \cellcolor[HTML]{ffd8b4}16 leaves (509 params)  \\  \midrule \multirow{2}{*}{ICCT-complete} & \cellcolor[HTML]{fff4ea}$1000.0\pm0.0$ & \cellcolor[HTML]{fff4ea}\textbf{300.5 $\pm$ 1.2} & \cellcolor[HTML]{fff3e8}\textbf{476.6 $\pm$ 3.1} & \cellcolor[HTML]{ffeddc}$120.7\pm0.5$ & \cellcolor[HTML]{ffeede}$1248.6\pm3.6$ & \cellcolor[HTML]{ffe3c9} $994.1\pm29.1$  \\   & \cellcolor[HTML]{fff4ea}2 leaves (13 params) & \cellcolor[HTML]{ffeddc}8 leaves (165 params) & \cellcolor[HTML]{ffe3c9}16 leaves (253 params) & \cellcolor[HTML]{ffd5ad}16 leaves (765 params) & \cellcolor[HTML]{ffc288}16 leaves (2189 params) &  \cellcolor[HTML]{ffd8b4}16 leaves (509 params)  \\  \midrule  \multirow{2}{*}{CDDT-controllers Crisp} & \cellcolor[HTML]{ff8710}$84.0\pm10.4$ & \cellcolor[HTML]{ffb065}$-126.6\pm53.5$ & \cellcolor[HTML]{ff850c}$-39826.4\pm21230.0$ & \cellcolor[HTML]{ffbd7e}$97.9\pm12.0$ & \cellcolor[HTML]{ff9228}$639.62\pm160.4$ &  \cellcolor[HTML]{ff7f00}$245.5\pm48.5$  \\   & \cellcolor[HTML]{fff4ea}2 leaves (13 params) & \cellcolor[HTML]{ffeddc}8 leaves (165 params) & \cellcolor[HTML]{ffe3c9}16 leaves (253 params) & \cellcolor[HTML]{ffd5ad}16 leaves (765 params) & \cellcolor[HTML]{ffc288}16 leaves (2189 params) & \cellcolor[HTML]{ffd8b4} 16 leaves (509 params)   \\  \specialrule{.2em}{.1em}{.1em} \multirow{2}{*}{MLP-Lower} & \cellcolor[HTML]{fff4ea}$1000.0\pm0.0$ & \cellcolor[HTML]{ffe8d3}$231.6\pm49.8$  & \cellcolor[HTML]{fff3e8}$474.7\pm5.8$ & \cellcolor[HTML]{fff0e3}\textbf{121.8 $\pm$ 0.6} & \cellcolor[HTML]{ff932a}$646.4\pm151.2$ & \cellcolor[HTML]{ffd1a6}$868.4\pm100.9$  \\   & \cellcolor[HTML]{ffefe1}79 params & \cellcolor[HTML]{fff0e3}110 params & \cellcolor[HTML]{ffeede} 127 params & \cellcolor[HTML]{fff0e3}151 params & \cellcolor[HTML]{fff2e5}221 params & \cellcolor[HTML]{fff2e5} 103 params  \\  \midrule \multirow{2}{*}{MLP-Upper} & \cellcolor[HTML]{fff4ea}$1000.0\pm0.0$ & \cellcolor[HTML]{fff2e5}$288.7\pm2.8$ & \cellcolor[HTML]{fff2e5}$467.9\pm8.5$ & \cellcolor[HTML]{fff0e3}$121.8\pm0.3$ & \cellcolor[HTML]{ffeddc}$1239.5\pm4.2$ &  \cellcolor[HTML]{ffeede}$1077.7\pm31.1$  \\   & \cellcolor[HTML]{ffebda} 121 params & \cellcolor[HTML]{ffe9d5}222 params & \cellcolor[HTML]{ffd7b2}407 params & \cellcolor[HTML]{ffd7b2}709 params & \cellcolor[HTML]{ffa854}3266 params &  \cellcolor[HTML]{ffb977}1021 params  \\  \midrule  \multirow{2}{*}{MLP-Max} & \cellcolor[HTML]{fff4ea}$1000.0\pm0.0$ & \cellcolor[HTML]{fff3e8} \textbf{298.5 $\pm$ 0.7} & \cellcolor[HTML]{fff4ea}\textbf{478.2 $\pm$ 6.7} & \cellcolor[HTML]{fff0e3}$121.7\pm0.4$ & \cellcolor[HTML]{ffca98}$1011.9\pm141.3$ & \cellcolor[HTML]{fff2e5}\textbf{1104.3 $\pm$ 9.4}   \\   & \cellcolor[HTML]{ff7f00} 67329 params & \cellcolor[HTML]{ff7f00}68610 params & \cellcolor[HTML]{ff7f00}69377 params & \cellcolor[HTML]{ff7f00} 77569 params & \cellcolor[HTML]{ff7f00}83458 params & \cellcolor[HTML]{ff7f00}73473 params  \\  \midrule  \multirow{2}{*}{CDDT} & \cellcolor[HTML]{fff4ea}\textbf{1000.0 $\pm$ 0.0} & \cellcolor[HTML]{ffe8d3}$226.4\pm44.5$ & \cellcolor[HTML]{fff2e5}$464.7\pm5.4$ & \cellcolor[HTML]{ffeddc}$120.9\pm0.5$ & \cellcolor[HTML]{ffeede}\textbf{1248.0 $\pm$ 6.4} &  \cellcolor[HTML]{ffe8d3}$1033.2\pm24.1$  \\   & \cellcolor[HTML]{fff4ea}2 leaves (8 params) & \cellcolor[HTML]{fff2e5} 8 leaves (86 params) & \cellcolor[HTML]{ffe5ce}16 leaves (226 params) & \cellcolor[HTML]{ffd8b4}16 leaves (706 params) & \cellcolor[HTML]{ffdec0}16 leaves (1036 params) &  \cellcolor[HTML]{ffdbb9}16 leaves (466 params)  \\  \midrule \multirow{2}{*}{CDDT-controllers} & \cellcolor[HTML]{fff4ea}$1000.0\pm0.0$ & \cellcolor[HTML]{fff2e5}$289.0\pm2.4$  & \cellcolor[HTML]{fff3e8}$469.7\pm11.1$ & \cellcolor[HTML]{ffeddc}$120.1\pm0.3$ & \cellcolor[HTML]{ffeede}$1243.8\pm3.6$ &  \cellcolor[HTML]{ffe5ce}$1010.9\pm25.7$  \\   & \cellcolor[HTML]{fff4ea}2 leaves (16 params) & \cellcolor[HTML]{ffe9d5} 8 leaves (214 params) & \cellcolor[HTML]{ffd6af}16 leaves (418 params) & \cellcolor[HTML]{ffb977}16 leaves (1410 params) & \cellcolor[HTML]{ffc48c}16 leaves (2092 params) & \cellcolor[HTML]{ffbf83}16 leaves (914 params)  \\   \specialrule{.2em}{.1em}{.1em} \end{tabular}

}
\label{tab:my-table}
\end{table*}

\section{Environments}
\label{sec:environments}
Here, we provide short descriptions across six domains used in our extensive evaluation. \textcolor{black}{We start with two common continuous control problems, Inverted Pendulum and Lunar Lander provided by OpenAI Gym \cite{brockman2016openai}. We then test across four autonomous driving scenarios: Lane-Keeping provided by \citet{highway-env} and Single-Lane Ring Network, Multi-Lane Ring Network, and Figure-8 Network all provided by the Flow deep reinforcement learning framework for mixed autonomy traffic scenarios \cite{Wu2017FlowAA}.} We provide additional details and depictions of each domain within the Appendix (Section \ref{sec:sup_environments}).

\noindent \textbf{Inverted Pendulum:} In Inverted Pendulum \cite{todorov2012mujoco}, a control policy must apply throttle \textcolor{black}{(ranging from +3 to move left to -3 to move right) to balance a pole. The observation includes the cart position, velocity, pole angle, and pole angular velocity.} 

\noindent \textbf{Lunar Lander:} In Lunar Lander \cite{parberry2017introduction,brockman2016openai}, a policy must throttle main engine and side engine thrusters for a lander to land on a specified landing pad.
\textcolor{black}{The observation is 8-dimensional including the lander's current position, linear velocity, tilt, angular velocity, and information about ground contact. The continuous action space is two dimensional for controlling the main engine thruster and side thrusters.}

\noindent\textbf{Lane-Keeping \cite{highway-env}:} A control policy must control a vehicle's steering angle to stay within a curving lane. 
\textcolor{black}{The observation is 12-dimensional, which consists of the vehicle's lateral position, heading, lateral speed, yaw rate, linear, lateral, and angular velocity, and the lane information. 
The action is the steering angle to control the vehicle.}

\noindent\textbf{Flow Single-Lane Ring Network \cite{Wu2017FlowAA}:} A control policy must apply acceleration commands to a vehicle agent to stabilize traffic flow consisting of 21 other human-driven (synthetic) vehicles. 
\textcolor{black}{The observation includes the world position on velocity of all vehicles.}

\noindent\textbf{Flow Multi-Lane Ring Network \cite{Wu2017FlowAA}:} A control policy must apply acceleration and lane-changing commands to an ego vehicle to stabilize the flow of noisy traffic flow across multiple lanes. \textcolor{black}{The observation includes the world position and velocity of all vehicles.}

\noindent\textbf{Flow Figure-8 Network \cite{Wu2017FlowAA}:} A control policy must apply acceleration to a vehicle to stabilize the flow in a Figure-8 network \textcolor{black}{(contains a section where the vehicles must cross paths at the center of the 8), requiring the policy to adapt its control input to create a stable flow through this section. The observation is the world position and velocity of all vehicles.}


\section{Results}
\label{sec:results}
In this section, we present the set of baselines we test our model against. Then, we report the results of our approach versus these baselines across the six continuous control domains, as shown in Table 1.
All presented results are across five random seeds and all differentiable frameworks are trained via SAC \cite{haarnoja2018soft}. Each tree-based framework is trained while maximizing performance and minimizing the complexity required to represent such a policy, thereby emphasizing interpretability. 
We release our codebase at \href{https://github.com/CORE-Robotics-Lab/ICCT}{\textcolor{blue}{https://github.com/CORE-Robotics-Lab/ICCT}}.
\subsection{Baselines}
We provide a list of baselines alongside abbreviations used for reference and brief definitions below. We compare against interpretable models, black-box models, and models that can be converted post-hoc into an interpretable form. \textcolor{black}{We also include the number of parameters\footnote{We only consider the active parameters involved during the deployment of the trained model. } for each method, shown in Table~\ref{tab:my-table}. We list the following notations for an easier understanding of the number of parameters. The number of leaf nodes is $n_l$ (the number of decision nodes is $n_l-1$). The dimension of the observation space is $m$. The number of active features within the leaf controllers is $e$. The dimension of the action space is $d_a$. The calculated number of parameters is denoted as $n_p$. Our approach, ICCT-$e$-feature, has a number of parameters of $n_p=3(n_l-1)+(2e+1)d_a n_l=(2ed_a+d_a+3)n_l-3$. }

\begin{itemize}[leftmargin=*]
    \item Continuous DDTs (CDDT): We translate the framework of \citet{Silva2021EncodingHD} to function with continuous action-spaces by modifying the leaf nodes to represent static probability distributions. \textcolor{black}{Here, $n_p=(m+2)(n_l-1)+d_a n_l=(d_a+m+2)n_l-m-2$. When converted into an interpretable form post-hoc, this approach is reported as CDDT-crisp which has a number of parameters: $n_p=3(n_l-1)+d_a n_l=(3+d_a)n_l-3$. }
    \item Continuous DDTs with controllers (CDDT-controllers): We modify CDDT leaf nodes to utilize linear controllers rather than static distributions. \textcolor{black}{Here, $n_p=(m+2)(n_l-1)+(m+1)d_a n_l=(md_a+d_a+m+2)n_l-m-2$. When converted into an interpretable form post-hoc, this approach is reported as CDDT-controllers Crisp that has $n_p=3(n_l-1)+(m+1)d_a n_l=(md_a+d_a+3)n_l-3$}. 
    \item ICCTs with static leaf distributions (ICCT-static): We modify the leaf architecture of our ICCTs to utilize static distributions for each leaf (i.e., set $e=0$). Comparing ICCT and ICCT-static displays the effectiveness of the addition of sparse linear sub-controllers. \textcolor{black}{Here, $n_p=3(n_l-1)+d_a n_l=(3+d_a)n_l-3$.}
    \item ICCT with complete linear sub-controllers (ICCT-complete): We allow the leaf controllers to maintain weights over all features \textcolor{black}{(no sparsity enforced, i.e., $e=m$}). Comparing ICCT-complete and CDDT-controllers displays the effectiveness of the proposed differentiable crispification procedure. \textcolor{black}{Here, $n_p=3(n_l-1)+(m+1)d_a n_l=(md_a+d_a+3)n_l-3$.}
    \item ICCT with L1-regularized controllers (ICCT-L1-sparse): We achieve sparsity via L1-regularization applied to ICCT-complete \textcolor{black}{rather than enforce sparsity directly via the \textsc{Enforce$\_$Controller$\_$Sparsity} procedure. While this baseline produces sparse sub-controllers, there are drawbacks limiting its interpretability. L1-regularization enforces weights to be near-zero rather than exactly zero. These small weights must be represented within decision-nodes and thus, the interpretability of the resulting model is limited.} \textcolor{black}{Here, $n_p=3(n_l-1)+(m+1)d_a n_l=(md_a+d_a+3)n_l-3$.}
    \item Multi-layer Perceptron (MLP): We maintain three variants of an MLP. The first (MLP-Max) contains a very large number of parameters, typically utilized in continuous control domains. The second (MLP-Upper) maintains approximately the same number of parameters of our ICCTs \textcolor{black}{with sparse leaf controllers} during training, including all inactive parameters after training (e.g., non-top feature weights in decision nodes). The last (MLP-Lower) maintains approximately the same number of \emph{active} parameters \textcolor{black}{as our ICCTs with sparse leaf controllers during evaluation. The number of parameters of MLP depends on the size of the network and we count all the weights and bias parameters but leave out all the optimizer parameters}.
    \item Decision Tree (DT): We train a DT via CART \cite{Breiman1983ClassificationAR} on state-action pairs generated from MLP-Max. \textcolor{black}{This baseline represents the distillation approach from a high-performance black-box policy to an interpretable model.} \textcolor{black}{Here, $n_p=2(n_l-1)+d_a n_l=(2+d_a)n_l-2$}. 
    \item DT w$\backslash$ \textcolor{black}{DAgger}: We utilize the \textcolor{black}{DAgger} imitation learning algorithm \cite{Ross2011ARO} to train a DT to mimic MLP-Max.
\end{itemize}
\subsection{Discussion}
We present the results of our trained policies in Table \ref{tab:my-table}. We provide the performance of each method alongside the associated complexity of each benchmark in Table \ref{tab:my-table} across three sections, with the top section representing interpretable approaches that maintain static distributions at their leaves, the middle section containing interpretable approaches that maintain linear controllers at their leaves, and the bottom section containing black-box methods.  

\textbf{Static Leaf Distributions (Top):} The frameworks of DT, DT w$\backslash$ \textcolor{black}{DAgger}, CDDT-Crisp, ICCT-static maintain similar representations and are equal in terms of interpretability given that the approaches have the same depth. \textcolor{black}{We see that across three of the six domains, ICCT-static is able to widely outperform both the DT and CDDT-Crisp models. In the remaining three domains, ICCT-static outperforms CDDT-Crisp by a large margin, and achieves competitive performance compared to DTs, even without accessing any superior expert policy.}


\textbf{Controller Leaf Distributions (Middle):} Here, we rank frameworks \textcolor{black}{(top-down)} by their relative interpretability. As the sparsity of the sub-controller decreases, the interpretability diminishes. We see that most approaches are able to achieve the maximum performance in the simple domain of Inverted Pendulum. However, CDDT-controllers-crisp encounters an inconsistency issue from the crispification procedure of \cite{Silva2021EncodingHD,Paleja2020InterpretableAP} and achieves very low performance. 
In regards to interpretability-performance tradeoff,
in Inverted Pendulum, we see that as sparsity increases within the sub-controller, a lower-depth ICCT can be used to achieve a equally high-performing policy. We note that across all domains, we do not find such a linear relationship. \textcolor{black}{We provide additional results within Section \ref{sec:interp-performance} that provide deeper insight into the interpretability-performance tradeoff.}

\textbf{Black-Box Approaches (Bottom):} MLP-based approaches and fuzzy DDTs are not interpretable. While the associated approaches perform well across many of the six domains, the lack of interpretability limits the utility of such frameworks in real-world applications \textcolor{black}{such as autonomous driving. We see that in half the domains, highly-parameterized architectures with over 65,000 parameters are required to learn effective policies (denoted by the dark orange shade).}

\noindent\textbf{Comparison Across All Approaches:} We see that across all continuous control domains, CDDT-Crisp and CDDT-controllers Crisp typically are the lowest-performing models. This displays the drawbacks of the crispification procedure of \cite{Silva2021EncodingHD,Paleja2020InterpretableAP} and the resultant performance inconsistency. Comparing our ICCTs to black-box models, we see that in all domains, we parity or outperform deep highly-parameterized models in performance while reducing the number of parameters required by orders of magnitude. In the difficult Multi-Lane Ring scenario, we see to we can outperform MLPs by 33$\%$ on average while achieving a $300$x-$600$x reduction in the number of policy parameters required.

Overall, we find extremely positive support for our Interpretable Continuous Control Trees, displaying the ability to at least parity black-box approaches while maintaining high interpretability. Our novel architecture and training procedure provide a strong step towards providing solutions for two grand challenges in interpretableML: (1) Optimizing sparse logical models such as DTs and (10) Interpretable RL.

\section{Qualitative Exposition of ICCT Interpretability}
\label{sec:qualitative}
\begin{figure}[t]
    \centering
    \includegraphics[width=0.43\textwidth]{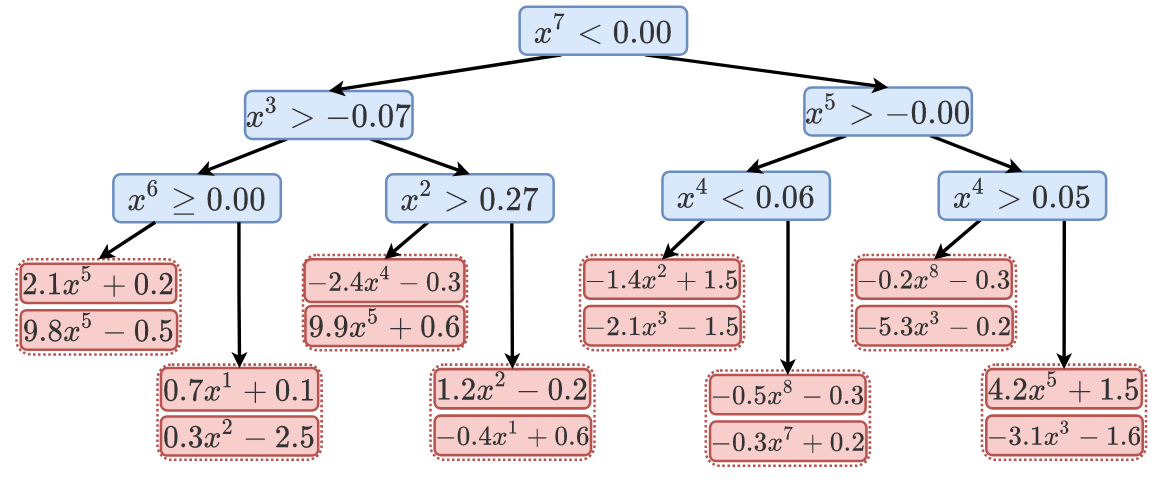}
    \caption{A Learned ICCT in Lunar Lander}
    \label{fig:icct_ll}
\end{figure}
\textcolor{black}{Here, we provide a display of the utility and interpretability of a learned ICCT model. In Figure \ref{fig:icct_ll}, we present our learned ICCT model in Lunar Lander, rounding each element to two decimal places for brevity. The displayed figure is an ICCT-1-feature model (i.e., only one active feature within the sparse sub-controller). The 8-dimensional input in Lunar Lander is composed of position ($x^1$,$x^2$), velocity ($x^3$,$x^4$), angle ($x^5$), angular velocity ($x^6$), left ($x^7$) and right ($x^8$) lander leg-to-ground contact. The action space is two-dimensional: the first (dictated by the top of each pair of the red-colored leaves) controls the main engine thrust, and the second (bottom) controls the net thrust for the side-facing engines. The tree can be interpreted as follows: \textcolor{black}{taking the 
leftmost path as an example, if the left leg is not touching the ground ($\leq0.00$ m), the horizontal velocity is greater than -0.07 m/s, and the angular velocity is greater than 0.00 rad/s, then the main engine action is $2.1 * \text{(the lander angle)} + 0.2$, and the side engine action is $9.8 * \text{(the lander angle)} - 0.5$}. Such a tree has several use cases: 1) An engineer/developer may pick certain edge cases and verify the behavior of the lander. Tree-based models are amerable to verification \cite{Vasic2019MoETMO}.
\emph{Furthermore, tree-based models similar to ICCTs can be verified in linear time \cite{Chen2019RobustnessVO}, while DNN verification is NP-complete \cite{Katz2017ReluplexAE}.}
2) An engineer can evaluate the decision-making in the tree and detect anomalies.} 
\textcolor{black}{Furthermore, there are hands-on use-cases of such a model, such as threshold editing (directly modifying nodes to increase affordances), etc.}

\begin{figure}[t]
\centering
    \begin{subfigure}[b]{0.47\textwidth}
    \includegraphics[width=0.94\textwidth, height = 5.2cm]{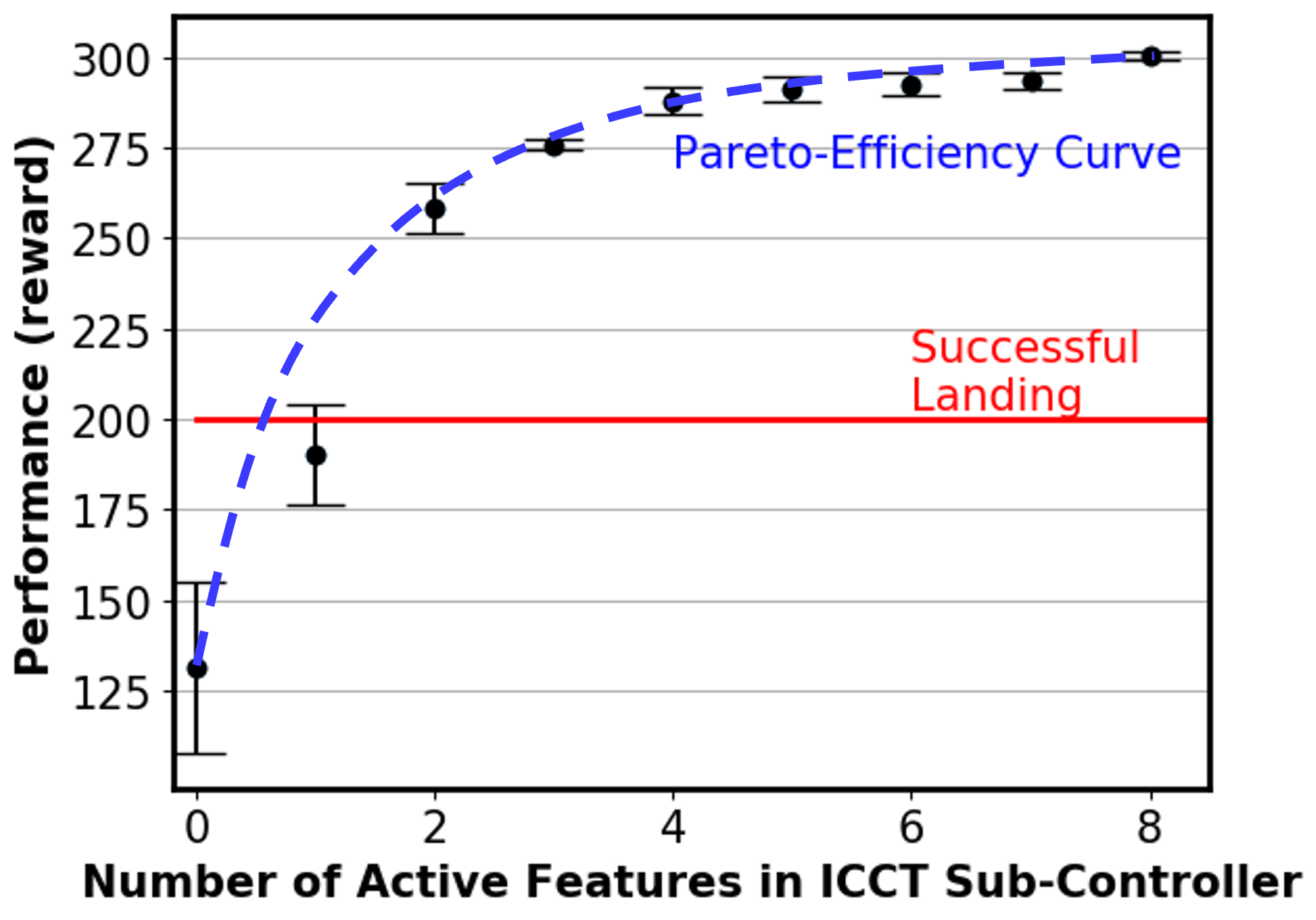}
    \caption{\textcolor{black}{Performance vs. Number of Controller Features}}
    \label{fig:pareto_sparse}
    \end{subfigure}
    ~~
    \begin{subfigure}[b]{0.45\textwidth}
    \includegraphics[width=0.89\textwidth,height = 5.2cm]{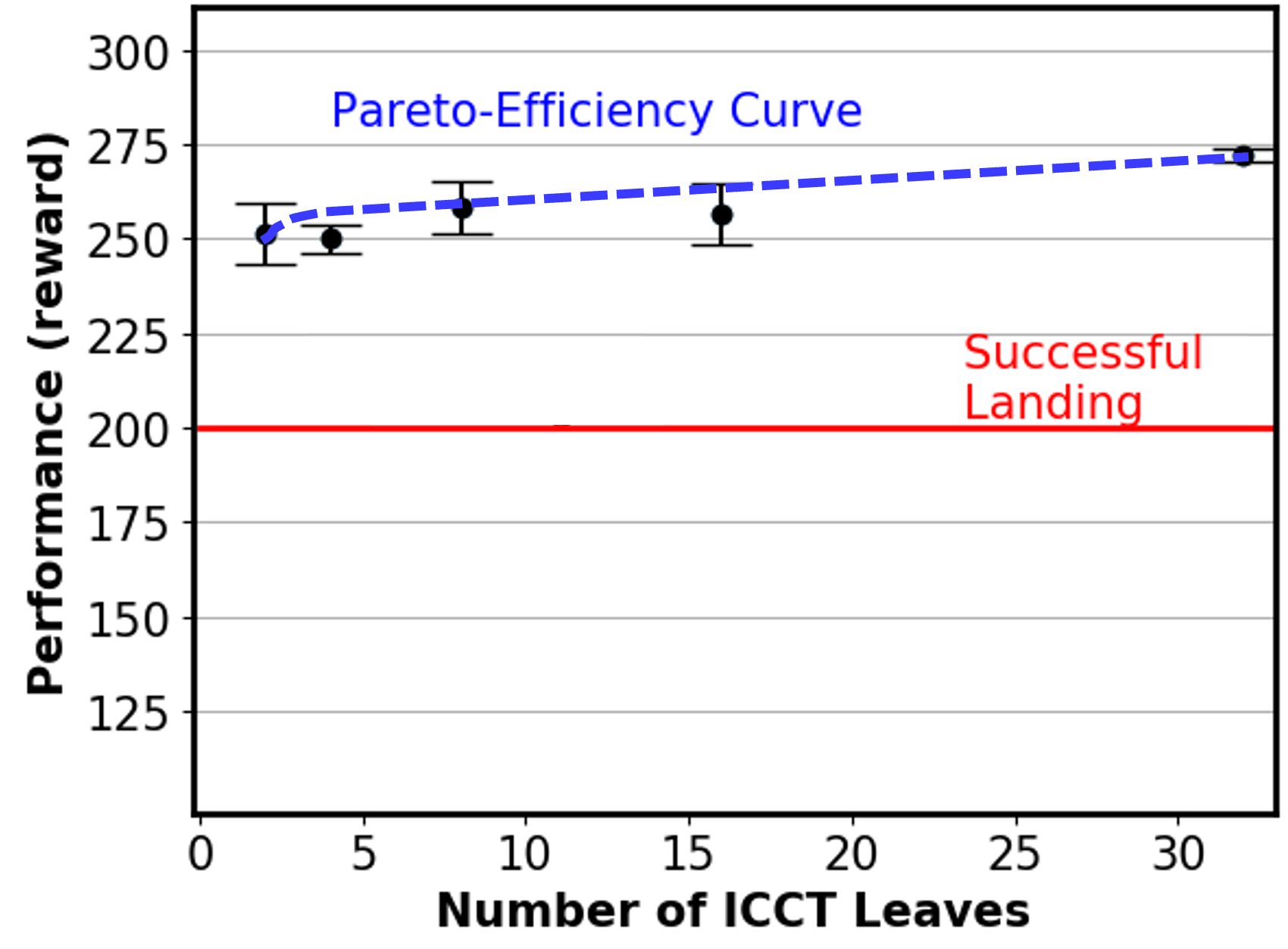}
    \caption{\textcolor{black}{Performance vs. Number of ICCT Leaves}}
    \label{fig:pareto_depth}
    \end{subfigure}
    
\caption{\color{black} In this figure, we display the interpretability-performance tradeoff of our ICCTs with respect to active features within our linear sub-controllers (Figure \ref{fig:pareto_sparse}) and tree depth (Figure \ref{fig:pareto_depth}). Within each figure, we display the Pareto-Efficiency Curve and denote the reward required for a successful lunar landing as defined by \cite{brockman2016openai}.}
\end{figure}

\section{Ablation: Interpretability-Performance Tradeoff}
\label{sec:interp-performance}
\textcolor{black}{Here, we provide an ablation study over how ICCT performance changes with respect to the number of active features within our linear sub-controllers and depth of the learned policies. \citet{lakkaraju} states that decision trees are interpretable because of their simplicity and that there is a cognitive limit on how complex a model can be while also being understandable. Accordingly, for our ICCTs to maximize interpretability, we emphasize the sparsity of our sub-controllers and attempt to minimize the depth of our ICCTs. Here, we present a deeper analysis by displaying the performance of our ICCTs while varying the number of active features, $e$, from ICCT-static to ICCT-complete (Figure \ref{fig:pareto_sparse}), and varying the number of leaves maintained within the ICCT from \textcolor{black}{$n_l=2$} to \textcolor{black}{$n_l=32$}. We conduct our ablation study within Lunar Lander.}

\begin{figure*}[h]
    \centering
    \includegraphics[width=\textwidth]{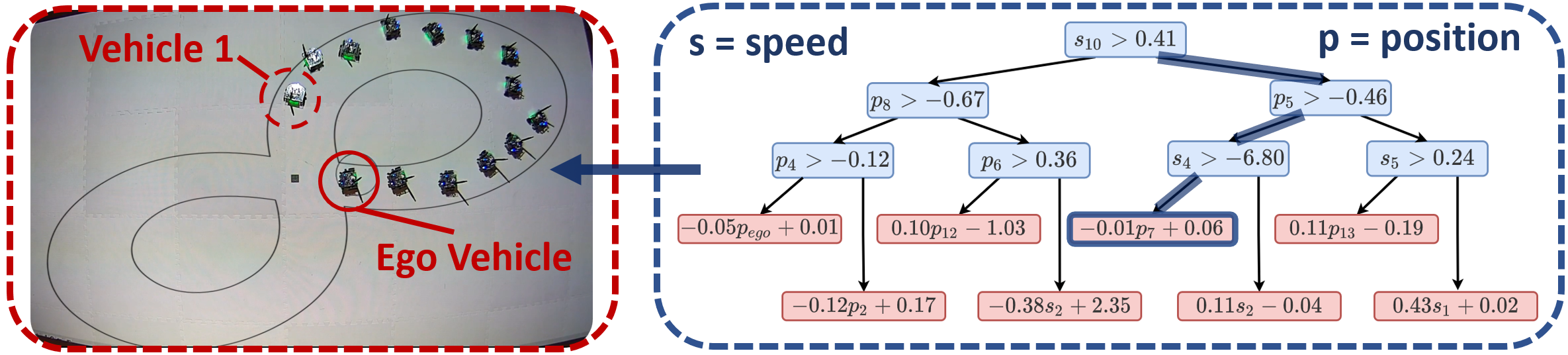}
    \caption{In this figure, we display our ICCTs controlling a vehicle in a 14-car physical robot demonstration within a Figure-8 traffic scenario. Active nodes and edges are highlighted by the right online visualization, where $s_i$ represents the speed of vehicle $i$, and $p_i$ represents the position of vehicle $i$. \textcolor{black}{We include a full video, including an enlarged display of our ICCT at \href{https://sites.google.com/view/icctree}{https://sites.google.com/view/icctree}.}}
    \label{fig:robotarium}
\end{figure*}
\textcolor{black}{In Figure \ref{fig:pareto_sparse}, we show how the performance of our ICCTs change as a function of active features in the Sub-Controller. Here, we fix the number of ICCT leaves to 8. We see that as the number of active features increase, the performance also increases. However, there is a tradeoff in interpretability. As above 200 reward is considered successful in this domain, a domain expert may determine a point on the Pareto-Efficiency curve that maximizes the interpretability-performance tradeoff.  In Figure \ref{fig:pareto_depth}, we show how the performance of our ICCTs change as a function of tree depth while fixing the number of active features in the ICCT sub-controller to two. We see a similar, albeit weaker, relationship between performance and interpretability. As model complexity increases, there is a slight gain in performance and a large decrease in interpretability. The Pareto-Efficiency curve provides insight into the interpretability-performance tradeoff for ICCT tree depth.}

\section{Physical Robot Demonstration}
\label{sec:robot}
\textcolor{black}{Here, we demonstrate our algorithm with physical robots in a 14-car figure-8 driving scenario and provide an online, easy-to-inspect visualization of our ICCTs, which controls the ego vehicle. We utilize the Robotarium, a remotely accessible swarm robotic research platform \cite{Wilson2020TheRG}, to demonstrate the learned ICCT policy. The demonstration displays the feasibility of ego vehicle behavior produced by our ICCT policy and provides an online visualization of our ICCTs. A frame taken from the demonstrated behavior is displayed in Figure \ref{fig:robotarium}. We provide a complete video of the demonstrated behavior and the online visualization of the control policy \textcolor{black}{at \href{https://sites.google.com/view/icctree}{https://sites.google.com/view/icctree}}.}

\section{Conclusion}
\label{sec:conclusion}
\textcolor{black}{In this work, we present a novel tree-based model for autonomous vehicle control.} Our Interpretable Continuous Control Trees (ICCTs) have competitive performance to that of deep neural networks across six continuous control domains, including four difficult autonomous driving scenarios, while maintaining high interpretability. The maintenance of both high performance and interpretability within an interpretable reinforcement learning architecture provides a paradigm that would be beneficial for the real-world deployment of autonomous systems. 

\section{Limitations and Future Work:} 
In planned work, we propose a deepening algorithm to dynamically grow ICCTs, allowing ICCTs to mitigate covariate shift and represent additional complexities when deployed to real-world settings. 
In the appendix, we present a sample of our deepening algorithm (Section \ref{sec:deepening}, which can currently be used to alleviate the need to set the number of ICCT leaves a priori. \textcolor{black}{We also plan to conduct a complete user-study, assessing if users can simulate ICCTs, verify ICCTs, and gain insight into the model's decision-making, similar to \cite{Silva2020OptimizationMF, Paleja2020InterpretableAP}.}

Our framework has several limitations. Continuous control outputs (e.g., predicting a steering angle) may not be interpretable to end-users and may require post-processing to enhance a user's understanding. Also, the relationship between controller sparsity, tree depth, and interpretability is not clear, making controller sparsity and tree depth difficult-to-define hyperparameters. \textcolor{black}{We also note that in more challenging environments, larger ICCTs may be required for representative power. However, in these cases, while ICCTs will be difficult to interpret by end-users due to their size, our ICCT policies can still be verified by experts and can be interpreted within tree sub-spaces.}

\section{Acknowledgments}
This work was supported by a gift award from the Ford Motor Company, NSF 1757401 (SURE Robotics), and a research grant from MIT Lincoln Laboratory (7000437192).

\bibliographystyle{plainnat}
\bibliography{references}

\newpage

\appendix
\subsection{Node Crispification Algorithm}
\label{sec:append_node_crisp}
\begin{figure}[h]
\centering
\includegraphics[width=0.36\textwidth]{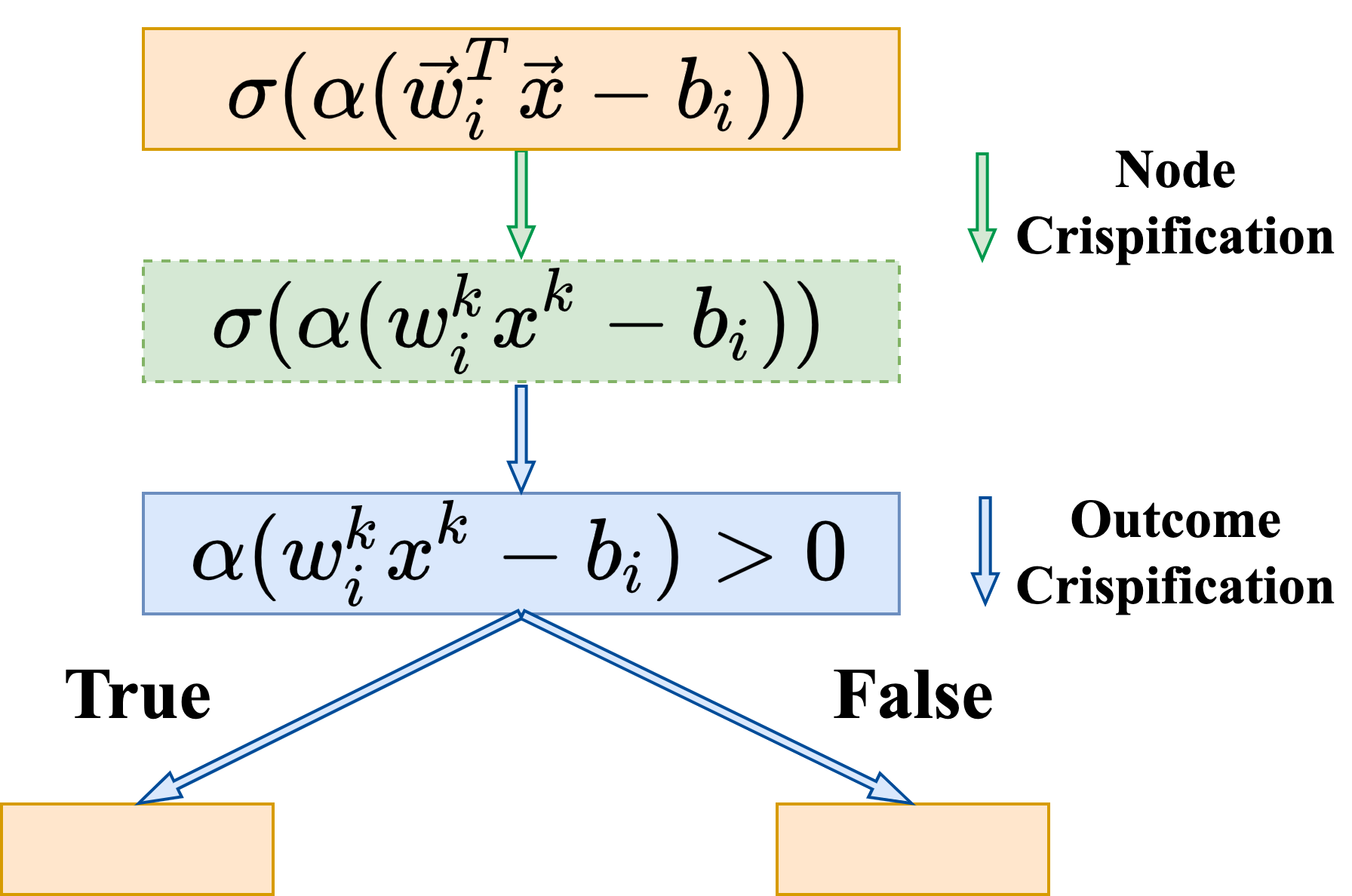}
\caption{\textcolor{black}{This figure displays the process of our proposed differentiable crispification, including node crispification (Algorithm \ref{alg:node_crisp}) and outcome crispification (Algorithm \ref{alg:outcome_crisp}). The node crispification sparsifies the weight vector $\vec{w_i}$ and chooses the most impactful feature. The outcome crispification helps make a ``hard'' decision instead of a ``soft'' decision to choose one branch. Both operations are differentiable.}}
\label{fig:node_outcome_crisp}
\end{figure}

In this section, we provide a description of node crispification, displayed in Algorithm \ref{alg:node_crisp} and termed \textsc{Node$\_$Crisp} in the main paper. We display the transformation performed by node crispification by the green arrow in Figure \ref{fig:node_outcome_crisp}. Node crispification recasts each decision node to split upon a single dimension of the input.

\begin{algorithm}[H]
\caption{\textcolor{black}{Node Crispification: \textsc{Node$\_$Crisp}$(\cdot)$}}
\label{alg:node_crisp}
\textbf{Input}: \small The original fuzzy decision node $\sigma(\alpha(\vec{w}^T_i \vec{x} - b_i))$, where $i$ is the decision node index, $\vec{w}_i=[w_i^1, w_i^2, ..., w_i^j, w_i^{j+1}, ..., w_i^m]^T$, and $m$ is the number of input features \\
\textbf{Output}: \small The intermediate decision node representation $\sigma(\alpha(w_i^k x^k - b_i))$ (see the green box in Figure \ref{fig:node_outcome_crisp}) 
\begin{algorithmic}[1] 
\STATE $\vec{z}_i=$ \textsc{diff\_argmax}$(|\vec{w}_i|)$ (\textsc{diff\_argmax}$(\cdot)$ displayed in Algorithm \ref{alg:argmax})
\STATE $\vec{w}_i' = \vec{z}_i\circ \vec{w}_i$
\STATE $\sigma(\alpha(w_i^k x^k - b_i)) = \sigma(\alpha(\vec{w}_i'^T \vec{x} - b_i))$ 
\end{algorithmic}
\end{algorithm}

\textcolor{black}{Node crispification takes as input the original fuzzy decision node, $\sigma(\alpha(\vec{w}^T_i \vec{x} - b_i))$, where \emph{all} input features are used in determining the output of decision node $i$. The output of this function is an intermediate decision node, $\sigma(\alpha(w_i^k x^k - b_i))$, where the output of decision node $i$ is only determined by \emph{a single }feature, $x^k$. To perform this transformation, in Line 1, we use the differentiable argument max function (in Algorithm \ref{alg:argmax}) to produce a one-hot vector, $\vec{z}_i$, with the element associated with the most impactful feature set to one and all other elements set to zero. In Line 2, we element-wise multiply the one-hot encoding, $\vec{z_i}$, by the original weights, $\vec{w}_i$, to produce a new set of weights with only one active weight, $\vec{w}_i'$. In Line 3, we show that by multiplying $\vec{x}$ by $\vec{w}_i'$, we can obtain the intermediate decision node $\sigma(\alpha(w_i^k x^k - b_i))$, where $k$ is the index of the most impactful feature (i.e., $k=\argmax_j(|w_i^j|)$).
}

\subsection{Outcome Crispfication Algorithm}
\label{sec:append_outcome_crisp}
\textcolor{black}{
In this section, we provide a description of outcome crispification, displayed in Algorithm \ref{alg:outcome_crisp} and termed \textsc{Outcome$\_$Crisp} in the main paper. We display the transformation performed by outcome crispification by the blue arrows in Figure \ref{fig:node_outcome_crisp}. Outcome crispification translates the outcome of a soft decision node to a hard decision node, resulting in a Boolean output from the decision node rather than a set of probabilities.
}

\begin{algorithm}[H]
\caption{\textcolor{black}{Outcome Crispfication: \textsc{Outcome$\_$Crisp}$(\cdot)$}}
\label{alg:outcome_crisp}
\textbf{Input}: \small The intermediate decision node $\sigma(\alpha(w_i^k x^k - b_i))$,  where $i$ is the decision node index, $k=\arg\max_j(|w_i^j|)$, and $w_i^j$ is the $j$th element in $\vec{w}_i$ \\
\textbf{Output}: \small Crisp decision node $\mathbbm{1}(\alpha(w_i^k x^k - b_i)>0)$ (see the blue box in Figure \ref{fig:node_outcome_crisp}) 
\begin{algorithmic}[1] 
\STATE $\vec{v}_i = [\alpha(w_i^k x^k - b_i), 0]$
\STATE $\vec{z}_i'=$ \textsc{diff\_argmax}$(\vec{v}_i)$ (\textsc{diff\_argmax}$(\cdot)$ displayed in Algorithm \ref{alg:argmax})
\STATE $\mathbbm{1}(\alpha(w_i^k x^k - b_i)>0) = \vec{z}_i'[0]$
\end{algorithmic}
\end{algorithm}

\textcolor{black}{Outcome crispification takes in the intermediate decision node $\sigma(\alpha(w_i^k x^k - b_i))$, which outputs the probability of branching left. The output of \textsc{Outcome$\_$Crisp} is the crisp decision node, $\mathbbm{1}(\alpha(w_i^k x^k - b_i)>0)$, a Boolean decision to trace down to the left branch OR right. In Line 1, we construct a soft vector representation of the decision node $i$'s output, $\vec{v}_i$, by concatenating $\alpha(w_i^k x^k - b_i)$ with a $0$. In Line 2, we use the differentiable argument max function (in Algorithm \ref{alg:argmax}) to produce a one-hot vector, $\vec{z}_i'$, where the first element represents the Boolean outcome of the decision node. In Line 3, we show that the output of the crisp decision node, $\mathbbm{1}(\alpha(w_i^k x^k - b_i)>0)$, can be obtained by choosing the first element of vector $\vec{z}_i'$ (we use bracket indexing notation here, starting with zero).} 


\subsection{\textcolor{black}{Differentiable Argument Max Function} for Differentiable Crispification}
\label{sec:diff_argmax}
In this section, we provide a description of the \textcolor{black}{differentiable argument max} function which is utilized in both decision node crispification and decision outcome crispification.

\begin{algorithm}[H]
\caption{Differentiable Argument Max Function for Crispification\textcolor{black}{: \textsc{diff\_argmax}$(\cdot)$}}
\label{alg:argmax}
\textbf{Input}: \small Logits $\vec{q}$ \\
\textbf{Output}: \small One-Hot Vector $\vec{h}$ 
\begin{algorithmic}[1] 
\STATE $\vec{h}_{soft} \leftarrow f(\vec{q})$
\STATE $\vec{h}_{hard} \leftarrow \textsc{one\_hot}(\textsc{argmax}(f(\vec{q})))$ \COMMENT{step 1 for $g(\cdot)$}
\STATE $\vec{h} = \vec{h}_{hard} +\vec{h}_{soft} - $\textsc{stop\_grad}($\vec{h}_{soft}$) \COMMENT{step 2 for $g(\cdot)$}
\end{algorithmic}
\end{algorithm}

\textcolor{black}{
Similar to \citep{Hafner2021MasteringAW}, we present a function call (in Algorithm \ref{alg:argmax}) that can be utilized to maintain gradients over a non-differentiable argument max operation. The function takes in a set of logits, $\vec{q}$, and applies a softmax operation, denoted by $f(\cdot)$, to output $\vec{h}_{soft}$, as shown in Line 1. In Line 2, the logits are transformed using an argument max followed by a one-hot procedure, causing the removal of gradient information, producing $\vec{h}_{hard}$. In Line 3, we combine $\vec{h}_{soft}$, $\vec{h}_{hard}$, and \textsc{stop\_grad}($\vec{h}_{soft}$) to output $\vec{h}$, \textcolor{black}{where \textsc{stop\_grad}($\cdot$) keeps the values and detaches the gradient data of $\vec{h}_{soft}$}. The outputted value of $\vec{h}$ is equal to that of $\vec{h}_{hard}$. However, the gradient maintained within $\vec{h}$ is associated with $\vec{h}_{soft}$. Automatic differentiation frameworks can then utilize the outputted term to perform backpropagation. Here, the operations in Line 2 and Line 3 compose function $g(\cdot)$ in Equation \ref{eq:crispification} and \ref{eq:decision_outcome_crispfication}.
}




\subsection{\textcolor{black}{Ablation: Differentiable Argument Max and Gumbel-Softmax}}
\label{sec:ablation_gumbel}

\begin{figure*}[t]
\centering
\includegraphics[width=0.78\textwidth]{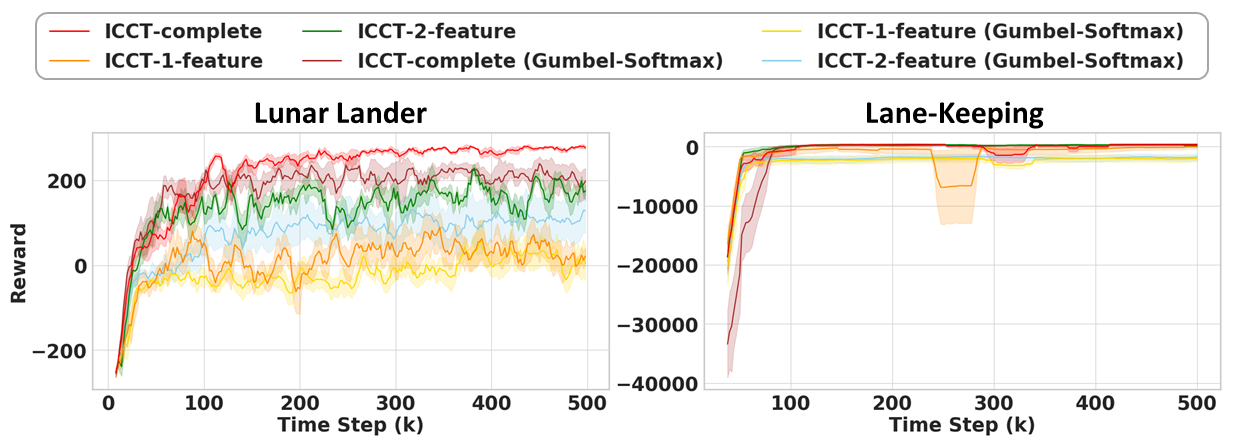}
\caption{This figure displays the average running rollout rewards of six methods for the ablation study during training. The results are averaged over 5 seeds, and the shadow region represents the standard error.}
\label{fig:ablation_gumbel}
\end{figure*}

\begin{table*}[t]
\centering 
\begin{tabular}{c c c} 
\hline
Method &  Lunar Lander & Lane-Keeping \\ [0.5ex] 
\hline %
ICCT-complete & $300.5\pm1.2$ & $476.6\pm3.1$ \\
ICCT-complete (Gumbel-Softmax) & $276.7\pm7.0$ & $412.6\pm31.3$ \\
ICCT-complete (Gumbel-Softmax, Crisp) & $239.0\pm18.9$ & $309.1\pm94.6$ \\
\hline
ICCT-1-feature & $190.1\pm13.7$ & $437.6\pm7.0$ \\
ICCT-1-feature (Gumbel-Softmax) & $113.2\pm43.1$ & $-853.4\pm333.2$  \\
ICCT-1-feature (Gumbel-Softmax, Crisp) & $-20.1\pm50.0$ & $-658.114\pm345.3$  \\
\hline
ICCT-2-feature & $258.4\pm7.0$ & $458.5\pm6.3$  \\
ICCT-2-feature (Gumbel-Softmax) & $161.7\pm54.8$ & $-560.6\pm251.6$  \\
ICCT-2-feature (Gumbel-Softmax, Crisp) & $62.3\pm82.2$ & $-945.0\pm331.0$  \\
\hline 
\end{tabular}
\caption{This table shows performance comparison between ICCTs utilizing \textcolor{black}{our proposed differentiable Argmax function (\textsc{diff\_argmax}$(\cdot)$ in Algorithm \ref{alg:argmax}), a variant of ICCTs utilizing the Gumbel-Softmax function, and a variant of crisp ICCTs utilizing the Gumbel-Softmax function}. Across each approach, we include ICCTs with fully parameterized sub-controllers (ICCT-complete) and sparse sub-controllers. We present our findings across Lunar Lander and Lane-Keeping.} 
\label{table:ablation_gumbel} 
\end{table*}
\normalsize

In this section, we provide an ablation study \textcolor{black}{on the differentiable operator used in ICCTs. Here, we substitute the Softmax function with a Gumbel-Softmax \citep{Jang2017CategoricalRW} function, a widely-used differentiable approximate sampling mechanism for categorical variables,} to perform decision node crispification, perform decision outcome crispification, and enforce sub-controller sparsity. \textcolor{black}{Changing ICCT to utilize the Gumbel-Softmax function as opposed to \textsc{diff\_argmax}$(\cdot)$ in Algorithm~\ref{alg:argmax} requires modifying the original Softmax function}, $f$, introduced by Equation \ref{eq:softmax}, to $f'$ as follows:
\begin{equation}
\label{eq:gumbel_softmax}
    f'(\vec{w_i})_k = \frac{\exp{\big(\frac{w_i^k + g_i^k}{\tau}\big)}}{\sum_j^m \exp{\big(\frac{w_i^j + g_i^j}{\tau}\big)}} 
\end{equation}
\textcolor{black}{Here, $\vec{w_i}$ is a $m$-dimensional vector, $[w_i^1, \cdots, w_i^m]^T$, and $\{g_i^j\}_{j=1}^m$ are i.i.d samples from a $\text{Gumbel}(0, 1)$ distribution \citep{Jang2017CategoricalRW}.} Here, we compare the performance of ICCT-complete, ICCT-1-feature, and ICCT-2-feature to their variants using Gumbel-Softmax in Lunar Lander and Lane-Keeping. All the methods and their corresponding variants are trained using the same hyperparameters.

From the results shown in Figure \ref{fig:ablation_gumbel} and Table \ref{table:ablation_gumbel},  we find that the addition of Gumbel noise reduces performance by a wide margin. Furthermore, comparing crisp ICCTs utilizing Gumbel-Softmax to ICCTs utilizing Gumbel-Softmax, we see that due to the sampling procedure within the Gumbel-Softmax, an inconsistency issue arises between non-crisp and crisp performance. \textcolor{black}{Such results support our design choice of the differentiable argument max function. }

\subsection{Universal Function Approximation}
\label{sec:ufa}
In this section, we provide a proof to show our ICCTs are universal function approximators, that is, can represent any decision surface given enough parameters. Our ICCT architecture consists of successive indicator functions, whose decision point lies among a single dimension of the feature space, followed by a linear controller to determine a continuous control output. For simplicity, we assume below that the leaf nodes contain static distributions. However, maintaining a linear controller at the leaves is more expressive and thus, the result below generalizes directly to ICCTs. 

The decision-making of our ICCTs can be decomposed as a sum of products. In Equation \ref{eq:decomp}, we display a computed output for a 4-leaf tree, where decision node outputs, $y_i$, are determined via Equation 1 of the main paper. Here, the sigmoid steepness, $\alpha$ is set to infinity (transforming the sigmoid function into an indicator function) resulting in hard decision points ($y_i \in \{0,1\}$. Equation \ref{eq:decomp} shows that the chosen action is determined by computation of probability of reaching a leaf, $y$, multiplied by static tree weights maintained at the distribution, $p$.
\begin{align}
\small
    \label{eq:decomp}
    ICCT(x) &= p_1 (y_1 *y_2) + p_2 (y_1 * (1-y_2)) \\ \nonumber
    & + p_3 ((1-y_1) * y_3) + p_4*((1-y_1) *(1-y_3))
\end{align}
\normalsize
Equation \ref{eq:decomp} can be directly simplified into the form of $G(x) = \sum_{j=1}^N p_j \sigma (w_j^T x + b_j)$, similar to Equation 1 in \cite{Cybenko1989ApproximationBS}. \cite{Cybenko1989ApproximationBS} demonstrates that finite combination of fixed, univariate functions can approximate any continuous function. The key difference between our architecture is that our univariate function is an indiator function rather than the commonly used sigmoid function. Below, we provide two lemmas to show that indicator functions fall within the space of univariate functions \cite{Cybenko1989ApproximationBS}.
\begin{lemma}
An indicator function is sigmoidal.
\end{lemma}
\textit{Proof:} This follows from the definition of sigmoidal: $\sigma(t)\to 1$ as $t \to \infty$ and $\sigma(t)\to 0$ as $t \to -\infty$.
\begin{lemma}
An indicator function is discriminatory.
\end{lemma}
\textit{Proof:} As an indicator function is bounded and measureable, by Lemma 1 of \cite{Cybenko1989ApproximationBS}, it is discriminatory.

\begin{theorem}
\label{thm:ufa}
Let $\sigma$ be any continuous discriminatory function. ICCTs are universal function approximators, that is, dense in the space of $C(I_n)$. 
In other words, there is a representation of ICCTs, $I(x)$, for which $|I(x) - f(x)| < \epsilon$ for all $x \in I_n$, for any function, f ($f \in C(I_n)$), where $C(I_n)$ denotes the codomain of an n-dimensional unit cube, $I_n$.
\end{theorem}
\textit{Proof:} As the propositional conditions hold for Theorem 1 in \cite{Cybenko1989ApproximationBS}, the result that ICCTs are dense in $C(I_n)$ directly follows. We note that as the indicator function 
jump-continuous, we refer readers to \cite{Selmic2002NeuralnetworkAO} whom extend UFA for $G(x) = \sum_{j=1}^N p_j \sigma (w_j^T x + b_j)$ to the case when $\sigma$ is jump-continuous.

\subsection{Learning Curves}
\begin{figure*}[htp]
    \centering
    \includegraphics[width=0.95\textwidth]{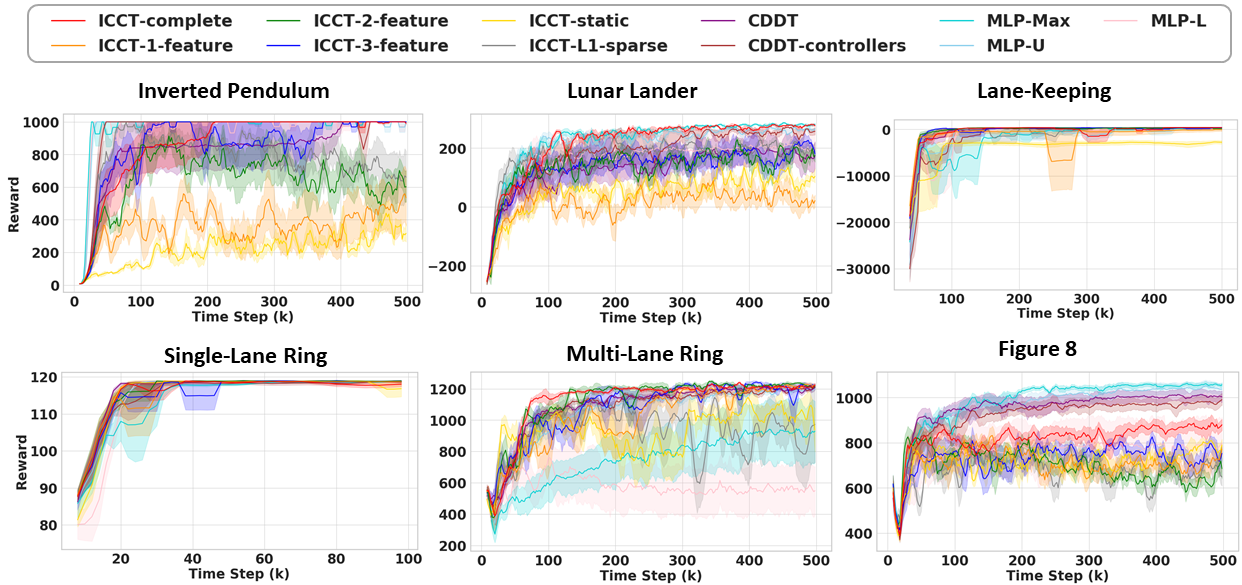}
    \caption{In this figure, we display the learning curves of eleven methods in six domains shown in the main paper. The rewards are rollout rewards throughout the training process. The curves in 5 domains except Lane-Keeping are smoothed by a sliding window of size 5, the curves in Lane-Keeping are smoothed by a sliding window of size 20. The shadow region depicts the standard error across 5 seeds.}
    \label{fig:learning_curves}
\end{figure*}
In Figure \ref{fig:learning_curves}, we display the learning curves of the eleven methods across six domains shown in Table 1 of the main paper. In general, ICCT-complete has competitive or better performance with regards to running-average rollout rewards and convergence rate, compared to MLP-Max, MLP-Upper and fuzzy DDTs in Inverted Pendulum, Lunar Lander, Lane-Keeping, Sing-Lane Ring, and Multi-Lane Ring, while maintaining interpretability. We also notice as the sparsity of the linear sub-controller increases, the performance of ICCT gradually drops. However, the interpretable approaches of ICCT-3-feature and ICCT-2-feature still have comparable or better performance with respect to MLP-Lower and ICCT-L1-sparse.

\subsection{Dynamic Depth}
\label{sec:deepening}
In this section, we present a dynamic deepening algorithm here drawing inspiration from \cite{Silva2021EncodingHD}. Allowing our ICCTs to automatically deepen and increase in complexity has several advantages.  In continuous control, it is typical that a deployed policy may encounter a covariate shift in a real-world setting \cite{Peng2018SimtoRealTO}. As such, our ICCT may need to change to account for features previously thought unimportant. Dynamic deepening would allow our ICCTs to mitigate the encountered covariate shift and represent additional complexities when deployed. Furthermore, the ability to deepen eliminates the need to set the tree depth a priori.

\begin{algorithm}[htp]
\caption{Dynamic Deepening Procedure}
\label{alg:deepening}
\textbf{Input}: \small Pretrained ICCT $P$, deepened ICCT $P_{deep}$, controller sparsity $e$ 
\begin{algorithmic}[1] 
\FOR{i epochs}
\STATE Collect trajectory rollouts, $\tau$, with $P$
\STATE $P,P_{deep}\leftarrow$\textsc{Network$\_$Update($P$,$P_{deep}$)} 
\STATE $\vec{H}, \vec{H}_{deep}\leftarrow$ \textsc{Calculate$\_$Leaf$\_$Entropies($P$,$P_{deep}$, $\tau$)}
\IF { $H(P_{deep}) + \epsilon< H(P)$ }
\STATE $P \leftarrow P_{deep}$
\STATE $P_{deep}\leftarrow$ \textsc{Deepen($P_{deep}$)}
\ENDIF

\ENDFOR
\end{algorithmic}
\end{algorithm}

Our proposed procedure for deepening is shown in Algorithm \ref{alg:deepening}. This procedure should be conducted after a ICCT model, $P$, has been pretrained with an initial dataset (e.g., with simulated data for transfer to the real world). At initialization of the dynamic deepening procedure, two ICCT models are maintained, a shallow pre-trained version, $P$, and deeper-by-one-depth version $P_{deep}$. Both models utilize the same controller sparsity $e$. The deepened ICCT is initialized so that the top-level weights match that of $P$, and lower-level weights are randomly initialized. During deployment, the pre-trained ICCT model, $P$ is utilized to collect trajectory rollouts, $\tau$, as shown in Line 1 of Algorithm \ref{alg:deepening}. Given these trajectory rollouts and associated rewards, both models are updated via gradient descent, as shown by the function \textsc{Network$\_$Update} in Line 3. $P$ is updated via Equation 2 in our main paper. As we do not have access to rollout trajectories for $P_{deep}$, the model update is simulated by utilizing the likelihood that $P_{deep}$ will take similar actions to $P$ given the states within $\tau$. In Line 4 of Algorithm \ref{alg:deepening} (\textsc{Calculate$\_$Leaf$\_$Entropies}), we calculate the entropy across each leaf within $P$ and $P_{deep}$. As our leaf nodes are input-parameterized (based on state), we utilize the sample of trajectories collected in Line 2 to estimate the leaf entropy. Here, $\vec{H}$ and $\vec{H}_{deep}$ represent a vector of leaf entropies. The deeper ICCT has more leaves (generated via the deepening procedure) and, thus, $\vec{H}_{deep}$ is a higher-dimensional vector. In Line 5 of Algorithm \ref{alg:deepening}, we compare the entropy of adjacent leaf nodes between $P$ and $P_{deep}$. For example, in the case where we have $P$ as a two-leaf tree and $P_{deep}$ as a four-leaf tree, if the entropy of the left leaf node of $P$ is at least $\epsilon$ greater than that of the combined entropy of the two left leaf nodes of $P_{deep}$, $P_{deep}$ has learned a leaf distribution that is more precise in representing high-performance control behavior. Thus, in Line 6 and 7 of Algorithm \ref{alg:deepening}, the shallow model, $P$, is updated the additional leaves of the deeper model, $P_{deep}$, and the deeper model, $P_{deep}$, is deepened by an additional level for each decision tree path that had lower entropy (determined in Line 5). This procedure continues for a set number of predefined epochs.

\begin{figure*}[h]
\centering
\includegraphics[width=0.83\textwidth]{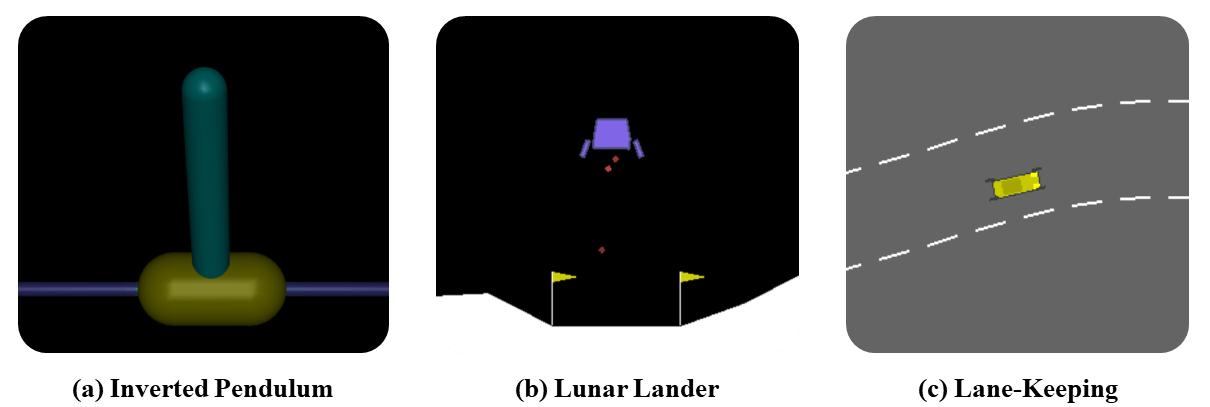}
\caption{This figure displays the first three environments we utilized including Inverted Pendulum, Lunar Lander, and Lane-Keeping.}
\label{fig:gym_env}
\end{figure*}

\begin{figure*}[h]
\centering
\includegraphics[width=0.83\textwidth]{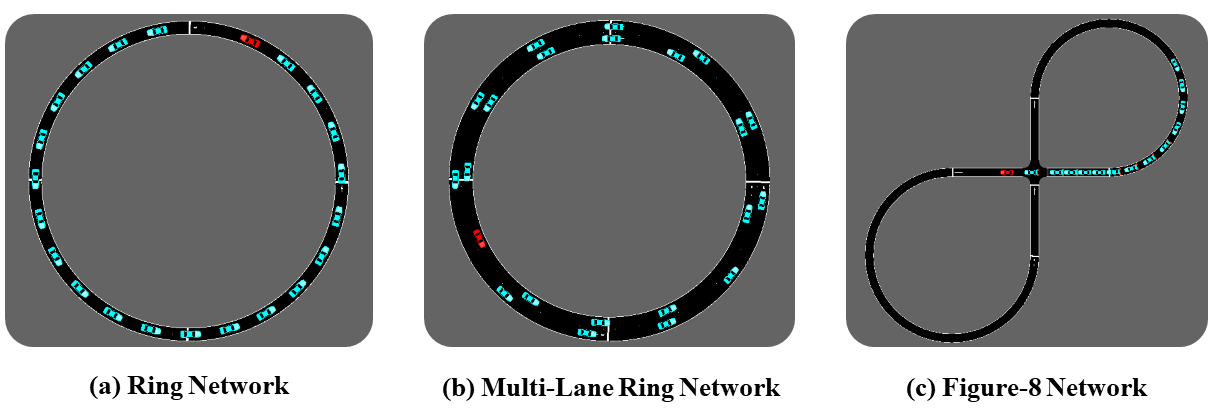}
\caption{In this figure, we display the configuration of each Flow network used in our experimentation.  In each image, the red vehicle is the autonomous agent that utilizes the learned control policy to stabilize traffic flow. Each cyan vehicle is a simulated human vehicle which contains a noisy acceleration behavior, the severity of which is defined by the user.}
\label{fig:sumo_env}
\end{figure*}

\subsection{Environments}
\label{sec:sup_environments}
We provide detailed descriptions for the two common continuous control problems: Inverted Pendulum and Lunar-Lander, and autonomous driving domains: Lane-Keeping and those from  Flow,
including Single-Lane Ring, Multi-Lane Ring, and Figure 8. 

\textbf{Inverted Pendulum:} Inverted Pendulum (Figure \ref{fig:gym_env}(a)) is provided by MujoCo \cite{todorov2012mujoco} and OpenAI Gym \cite{brockman2016openai}. 
The observation includes the cart position, velocity, pole angle, and pole angular velocity. 
The goal is to apply a force to the cart to balance a pole on it and prevent the pole from falling.
A plus one reward is provided at each timestep provided if the pole keeps upright at each time step.

\textbf{Lunar Lander:} Lunar Lander (Figure \ref{fig:gym_env}(b)) is a game provided by Box2D \cite{parberry2017introduction} and OpenAI  Gym. 
The goal is to land a lunar lander as close to a landing pad between the flags. 
The observation is 8-dimensional including the lander's current position, linear velocity, tilt, angular velocity, and information about ground contact. 
The continuous action space is two dimensional for controlling the main engine thruster and side thrusters.
At each timestep, the lander reward is determined by a proximity to the landing pad, whether each leg is touching the landing pad, and a fuel cost. The episode ends if the lander crashes and a terminal reward of -100 is provided. If the lander successfully lands, a terminal reward of 100 is provided. 

\textbf{Lane-Keeping:} Lane-Keeping (Figure \ref{fig:gym_env}(c))
is a domain with continuous actions within highway-env \cite{highway-env}. 
The observation is 12-dimensional, which consists of the vehicle's lateral position, heading, lateral speed, yaw rate, linear, lateral, and angular velocity, and the lane information. 
The action is the steering angle to control the vehicle. At each time step, +1 reward will be provided if the vehicle can keep in the center of the lane. The reward decreases as the vehicle drives away from the lane center.  The terminal condition within this domain is a maximum timestep of 500.

\textbf{Flow Domains \cite{Wu2017FlowAA}:} Each of the following domains are custom continuous actions domains provided within the Flow deep reinforcement learning framework for mixed autonomy traffic scenarios. Flow utilizes the SUMO traffic simulator, which allows for both autonomous agents and simulated human agents. The simulated human agents are constructed by adjusting the noise factor present within their acceleration and deceleration control. Stabilization of traffic is defined as the average velocity of all vehicles approaching a set value for velocity. The general observation includes the global positions and velocity of all vehicles in the network.  These are normalized based upon the length and max velocity for the network, which is defined by the user, and then combined into a single 1-Dimensional array with a length that is double the number of vehicles present. 
The reward is measured by how closely the network’s average velocity matches the user-defined average velocity.  The episode is terminated if any collision between two vehicles is detected, or preset time steps are executed.

\textbf{Flow Single-Lane Ring Network:} The Single-Lane Ring Network (Figure \ref{fig:sumo_env}(a)) is a ring road network, with the objective being to stabilize the flow of all traffic within the network. A control policy must apply acceleration commands to an autonomous agent in order to stabilize the flow. There are 21 human vehicles and 1 ego vehicle, with a maximum time step of 750.

\textbf{Flow Multi-Lane Ring Network:} The Multi-Lane Ring network (Figure \ref{fig:sumo_env}(b)) is a complex ring road network consisting of multiple lanes of traffic, with the objective being to stabilize the flow of all traffic within the network. The observation for this network includes the general observation for Flow networks, as well as the current lane index for each vehicle. The network holds 21 human vehicle and 1 ego vehicle.
A control policy must apply acceleration commands and lane changing commands (a continuous value from -1 to 1) to the ego vehicle in order to stabilize the flow of traffic between both lanes. The maximum step in this domain is set to 1500.

\textbf{Flow Figure-8 Network:} The Figure-8 Network (Figure \ref{fig:sumo_env}(c)) is a complex Flow domain as it contains a road section where the vehicles will cross-over, which is in the center of the figure-8. The control policy must consider this congestion point when controlling the ego vehicle, applying acceleration commands to the autonomous agent in order to stabilize the flow. The network holds 13 human vehicles besides the ego vehicle. Here the maximum time step is 1500.

\subsection{Hyperparameters}
All methods are trained using SAC with the same structure of critic network, which consists of an MLP with 2 hidden layers of 256 units. The buffer size is always set to 1000000, and the discount factor $\gamma$ is always set to 0.99. In the Single-Lane Ring network, the training steps are 100000, otherwise the training steps are 500000 for all the methods. The soft update coefficient, $\tau$, is set to 0.005 for ICCT-static in Inverted Pendulum,  and set to 0.01 in all other domains. We display the learning rates, batch sizes, and network sizes used for the methods discussed in the paper across each domain in Tables \ref{table:hyper-ip}-\ref{table:hyper-fig8}.

\begin{table}[htp]
\caption{Hyperparameters in Inverted Pendulum} 
\centering 
\scriptsize
\begin{tabular}{c c c c} 
\hline
Hyperparameter &  Learning Rate & Batch Size & Network Size\\ [0.5ex] 
\hline %
ICCT-complete & \num{3e-4} & 1024 & 2 leaves  \\
ICCT-1-feature & \num{6e-4} & 1024 & 8 leaves \\
ICCT-2-feature & \num{5e-4} & 1024 & 4 leaves \\
ICCT-3-feature & \num{5e-4} & 1024 & 2 leaves \\
ICCT-static & \num{5e-4} & 1024 & 32 leaves  \\
ICCT-L1-sparse & \num{3e-4} & 256 & 4 leaves \\
CDDT & \num{3e-4} & 1024 & 2 leaves \\
CDDT-controllers & \num{3e-4} & 1024 & 2 leaves \\
MLP-Max & \num{3e-4} & 1024 & [256, 256] \\
MLP-U & \num{3e-4} & 1024 & [8, 8] \\
MLP-L & \num{3e-4} & 1024 & [6, 6] \\
\hline 
\end{tabular}
\label{table:hyper-ip} 
\end{table}
\normalsize

\begin{table}[htp]
\caption{Hyperparameters in Lunar Lander} 
\centering 
\tiny
\begin{tabular}{c c c c c} 
\hline
Hyperparameter &  Actor Learning Rate &  Critic Learning Rate  & Batch Size & Network Size\\ [0.5ex] 
\hline %
ICCT-complete & \num{5e-4} & \num{5e-4} & 256 & 8 leaves  \\
ICCT-1-feature & \num{5e-4} & \num{5e-4} & 256 & 8 leaves \\
ICCT-2-feature & \num{5e-4} & \num{5e-4} & 256 & 8 leaves \\
ICCT-3-feature & \num{5e-4} & \num{5e-4} & 256 & 8 leaves \\
ICCT-static & \num{5e-4} & \num{3e-4} & 256 & 32 leaves  \\
ICCT-L1-sparse & \num{5e-4} & \num{5e-4} & 256 & 8 leaves \\
CDDT & \num{5e-4} & \num{3e-4} & 256 & 8 leaves \\
CDDT-controllers & \num{5e-4} & \num{3e-4} & 256 & 8 leaves \\
MLP-Max & \num{3e-4} & \num{3e-4} & 256 & [256, 256] \\
MLP-U & \num{3e-4} & \num{3e-4} & 256 & [10, 10] \\
MLP-L & \num{3e-4} & \num{3e-4} & 256 & [6, 6] \\
\hline 
\end{tabular}
\label{table:hyper-ll} 
\end{table}
\normalsize

\begin{table}[htp]
\caption{Hyperparameters in Lane-Keeping} 
\centering 
\scriptsize
\begin{tabular}{c c c c} 
\hline
Hyperparameter &  Learning Rate & Batch Size & Network Size\\ [0.5ex] 
\hline %
ICCT-complete & \num{3e-4} & 1024 & 16 leaves  \\
ICCT-1-feature & \num{3e-4} & 1024 & 16 leaves \\
ICCT-2-feature & \num{3e-4} & 1024 & 16 leaves \\
ICCT-3-feature & \num{3e-4} & 1024 & 16 leaves \\
ICCT-static & \num{2e-4} & 1024 & 16 leaves  \\
ICCT-L1-sparse & \num{3e-4} & 1024 & 16 leaves \\
CDDT & \num{3e-4} & 256 & 16 leaves \\
CDDT-controllers & \num{3e-4} & 512 & 16 leaves \\
MLP-Max & \num{3e-4} & 256 & [256, 256] \\
MLP-U & \num{3e-4} & 256 & [14, 14] \\
MLP-L & \num{3e-4} & 256 & [6, 6] \\
\hline 
\end{tabular}
\label{table:hyper-lk} 
\end{table}
\normalsize

\begin{table}[htp]
\caption{Hyperparameters in Sing-Lane Ring Network} 
\centering 
\scriptsize
\begin{tabular}{c c c c} 
\hline
Hyperparameter &  Learning Rate & Batch Size & Network Size\\ [0.5ex] 
\hline %
ICCT-complete & \num{5e-4} & 1024 & 16 leaves  \\
ICCT-1-feature & \num{5e-4} & 1024 & 16 leaves \\
ICCT-2-feature & \num{5e-4} & 1024 & 16 leaves \\
ICCT-3-feature & \num{5e-4} & 1024 & 16 leaves \\
ICCT-static & \num{5e-4} & 1024 & 16 leaves  \\
ICCT-L1-sparse & \num{5e-4} & 1024 & 16 leaves \\
CDDT & \num{5e-4} & 1024 & 16 leaves \\
CDDT-controllers & \num{5e-4} & 1024 & 16 leaves \\
MLP-Max & \num{3e-4} & 1024 & [256, 256] \\
MLP-U & \num{3e-4} & 1024 & [12, 12] \\
MLP-L & \num{3e-4} & 1024 & [3, 3] \\
\hline 
\end{tabular}
\label{table:hyper-ring-accel} 
\end{table}
\normalsize

\begin{table}[htp]
\caption{Hyperparameters in Multi-Lane Ring Network} 
\centering 
\scriptsize
\begin{tabular}{c c c c} 
\hline
Hyperparameter &  Learning Rate & Batch Size & Network Size\\ [0.5ex] 
\hline %
ICCT-complete & \num{5e-4} & 1024 & 16 leaves  \\
ICCT-1-feature & \num{5e-4} & 1024 & 16 leaves \\
ICCT-2-feature & \num{6e-4} & 1024 & 16 leaves \\
ICCT-3-feature & \num{5e-4} & 1024 & 16 leaves \\
ICCT-static & \num{5e-4} & 1024 & 16 leaves  \\
ICCT-L1-sparse & \num{5e-4} & 1024 & 16 leaves \\
CDDT & \num{5e-4} & 1024 & 16 leaves \\
CDDT-controllers & \num{5e-4} & 1024 & 16 leaves \\
MLP-Max & \num{5e-4} & 1024 & [256, 256] \\
MLP-U & \num{5e-4} & 1024 & [32, 32] \\
MLP-L & \num{5e-4} & 1024 & [3, 3] \\
\hline 
\end{tabular}
\label{table:hyper-ring-lc} 
\end{table}
\normalsize

\begin{table}[htp]
\caption{Hyperparameters in Figure-8 Network} 
\centering 
\scriptsize
\begin{tabular}{c c c c} 
\hline
Hyperparameter &  Learning Rate & Batch Size & Network Size\\ [0.5ex] 
\hline %
ICCT-complete & \num{5.5e-4} & 1024 & 16 leaves  \\
ICCT-1-feature & \num{6e-4} & 1024 & 16 leaves \\
ICCT-2-feature & \num{6e-4} & 1024 & 16 leaves \\
ICCT-3-feature & \num{7e-4} & 1024 & 16 leaves \\
ICCT-static & \num{5e-4} & 1024 & 16 leaves  \\
ICCT-L1-sparse & \num{5e-4} & 1024 & 16 leaves \\
CDDT & \num{5e-4} & 1024 & 16 leaves \\
CDDT-controllers & \num{5e-4} & 1024 & 16 leaves \\
MLP-Max & \num{5e-4} & 1024 & [256, 256] \\
MLP-U & \num{5e-4} & 1024 & [20, 20] \\
MLP-L & \num{5e-4} & 1024 & [3, 3] \\
\hline 
\end{tabular}
\label{table:hyper-fig8} 
\end{table}
\normalsize

\end{document}